\documentclass{article}

\PassOptionsToPackage{numbers, compress}{natbib}

\usepackage[preprint]{neurips_2024}

\usepackage[utf8]{inputenc} 
\usepackage[T1]{fontenc}    
\usepackage{hyperref}       
\usepackage{url}            
\usepackage{booktabs}       
\usepackage{amsfonts}       
\usepackage{nicefrac}       
\usepackage{microtype}      
\usepackage{xcolor}         

\usepackage{caption}
\usepackage{subcaption}
\usepackage{amsmath}
\usepackage{graphicx}
\usepackage{algorithm}
\usepackage{algorithmic}
\usepackage{amsthm}
\usepackage{amssymb}
\usepackage{thm-restate}
\usepackage{wrapfig}
\usepackage{makecell}

\newtheorem{theorem}{Theorem}[section]

\newtheorem{lemma}[theorem]{Lemma}

\title{Spectral-Risk Safe Reinforcement Learning \\ with Convergence Guarantees}

\author{%
  Dohyeong Kim$^1$ \quad Taehyun Cho$^1$ \quad Seungyub Han$^1$ \\
  \textbf{Hojun Chung}$^1$ \quad \textbf{Kyungjae Lee}$^{2*}$ \quad \textbf{Songhwai Oh}$^{1\!}$\thanks{Corresponding authors: kyungjae.lee@ai.cau.ac.kr, songhwai@snu.ac.kr.} \\
  $^{1}$Dep. of Electrical and Computer Engineering, Seoul National University \\
  $^2$Artificial Intelligence Graduate School, Chung-Ang University \\
}

\begin{document}

\maketitle

\begin{abstract}
The field of risk-constrained reinforcement learning (RCRL) has been developed to effectively reduce the likelihood of worst-case scenarios by explicitly handling risk-measure-based constraints.
However, the nonlinearity of risk measures makes it challenging to achieve convergence and optimality.
To overcome the difficulties posed by the nonlinearity, we propose a spectral risk measure-constrained RL algorithm, \emph{spectral-risk-constrained policy optimization (SRCPO)}, a bilevel optimization approach that utilizes the duality of spectral risk measures.
In the bilevel optimization structure, the outer problem involves optimizing dual variables derived from the risk measures, while the inner problem involves finding an optimal policy given these dual variables.
The proposed method, to the best of our knowledge, is the first to guarantee convergence to an optimum in the tabular setting.
Furthermore, the proposed method has been evaluated on continuous control tasks and showed the best performance among other RCRL algorithms satisfying the constraints.
\end{abstract}

\section{Introduction}


Safe reinforcement learning (Safe RL) has been extensively researched \citep{achiam2017constrained, kim2023sdac, xu2021crpo}, particularly for its applications in safety-critical areas such as robotic manipulation \citep{adjei2024safe, liu2024safe} and legged robot locomotion \citep{yang2022safe}. 
In safe RL, constraints are often defined by the expectation of the discounted sum of costs \citep{achiam2017constrained, xu2021crpo}, where cost functions are defined to capture safety signals. 
These expectation-based constraints can leverage existing RL techniques since the value functions for costs follow the Bellman equation.
However, they inadequately reduce the likelihood of significant costs, which can result in worst-case outcomes, because expectations do not capture information on the tail distribution.
In real-world applications, avoiding such worst-case scenarios is particularly crucial, as they can lead to irrecoverable damage.

In order to avoid worst-case scenarios, various safe RL methods \citep{chow2017risk, yang2023wcsac} attempt to define constraints using risk measures, such as conditional value at risk (CVaR) \citep{rockafellar2000optimization} and its generalized form, spectral risk measures \citep{acerbi2002spectral}.
\citet{yang2023wcsac} and \citet{zhang2022sdpo} have employed a distributional critic to estimate the risk constraints and calculate policy gradients.
\citet{zhang2024cvar} and \citet{chow2017risk} have introduced auxiliary variables to estimate CVaR.
These approaches have effectively prevented worst-case outcomes by implementing risk constraints. 
However, the nonlinearity of risk measures makes it difficult to develop methods that ensure optimality.
To the best of our knowledge, even for well-known risk measures such as CVaR, existing methods only guarantee local convergence \citep{chow2017risk, zhang2024cvar}.
This highlights the need for a method that ensures convergence to a global optimum.

In this paper, we introduce a spectral risk measure-constrained RL (RCRL) algorithm called \emph{spectral-risk-constrained policy optimization (SRCPO)}.
Spectral risk measures, including CVaR, have a dual form expressed as the expectation over a random variable \citep{bauerle2021minimizing}.
Leveraging this duality, we propose a bilevel optimization framework designed to alleviate the challenges caused by the nonlinearity of risk measures. 
This framework consists of two levels: the \emph{outer problem}, which involves optimizing a dual variable that originates from the dual form of spectral risk measures, and the \emph{inner problem}, which aims to find an optimal policy for a given dual variable.
For the inner problem, we define novel risk value functions that exhibit linearity in performance difference and propose a policy gradient method that ensures convergence to an optimal policy in tabular settings. 
The outer problem is potentially non-convex, so gradient-based optimization techniques are not suitable for global convergence.
Instead, we propose a method to model and update a distribution over the dual variable, which also ensures finding an optimal dual variable in tabular settings.
As a result, to the best of our knowledge, we present the first algorithm that guarantees convergence to an optimum for the RCRL problem.
Moreover, the proposed method has been evaluated on continuous control tasks with both single and multiple constraints, demonstrating the highest reward performance among other RCRL algorithms satisfying constraints in all tasks.
Our contributions are summarized as follows:
\begin{itemize}
    \item We propose a novel bilevel optimization framework for RCRL, effectively addressing the complexities caused by the nonlinearity of risk measures.
    \item The proposed method demonstrates convergence and optimality in tabular settings.
    \item The proposed method has been evaluated on continuous control tasks with both single and multiple constraints, consistently outperforming existing RCRL algorithms.

\end{itemize}

\section{Related Work}



RCRL algorithms can be classified according to how constraints are handled and how risks are estimated.
There are two main approaches to handling constraints: the Lagrangian \citep{ying2022cppo, yang2023wcsac, zhang2024cvar} and the primal \citep{kim2023sdac, zhang2022sdpo} approaches. 
The Lagrangian approach deals with dual problems by jointly updating dual variables and policies, which can be easily integrated into the existing RL framework. 
In contrast, the primal approach directly tackles the original problem by constructing a subproblem according to the current policy and solving it.
Subsequently, risk estimation can be categorized into three types.
The first type \citep{ying2022cppo, zhang2024cvar} uses an auxiliary variable to precisely estimate CVaR through its dual form. 
Another one \citep{yang2023wcsac, zhang2022sdpo} approximates risks as the expectation of the state-wise risks calculated using a distributional critic \citep{bellemare2017c51}. 
The third one \citep{kim2023sdac} approximates CVaR by assuming that cost returns follow Gaussians. 
Despite these diverse methods, they only guarantee convergence to local optima.
In the field of risk-sensitive RL, \citet{bauerle2021minimizing} have proposed an algorithm for spectral risk measures that ensures convergence to an optimum.
They introduced a bilevel optimization approach, like our method, but did not provide a method for solving the outer problem.
Furthermore, their approach to the inner problem is value-based, making it complicated to extend to RCRL, which typically requires a policy gradient approach.


\section{Backgrounds}

\subsection{Constrained Markov Decision Processes}

A constrained Markov decision process (CMDP) is defined as $\langle S, A, P, \rho, \gamma, R, \{C_i\}_{i=1}^N \rangle$, where $S$ is a state space, $A$ is an action space, $P$ is a transition model, $\rho$ is an initial state distribution, $\gamma$ is a discount factor, $R: S \times A \times S \mapsto [-R_\mathrm{max}, R_\mathrm{max}]$ is a reward function, $C_i: S \times A \times S \mapsto [0, C_\mathrm{max}]$ is a cost function, and $N$ is the number of cost functions.
A policy $\pi: S \mapsto \mathcal{P}(A)$ is defined as a mapping from the state space to the action distribution.
Then, the state distribution given $\pi$ is defined as $d_\rho^\pi(s) := (1-\gamma)\sum_{t=0}^\infty \gamma^t \mathbb{P}(s_t =s)$, where $s_0 \sim \rho$, $a_t \sim \pi(\cdot|s_t)$, and $s_{t+1} \sim P(\cdot|s_t, a_t)$ for $\forall t$.
The state-action return of $\pi$ is defined as:
\begin{equation*}
Z_R^\pi(s, a) := \sum_{t=0}^\infty \gamma^t R(s_t, a_t, s_{t+1}) \; \Big| \; s_0=s, \; a_0 = a, \; a_t \sim \pi(\cdot|s_t), \; s_{t+1} \sim P(\cdot|s_t, a_t) \; \forall t.
\end{equation*}
The state return and the return are defined, respectively, as:
\begin{equation*}
Y_R^\pi(s) := Z_R^\pi(s, a) \; \big| \; a \sim \pi(\cdot|s), \quad G_R^\pi := Y_R^\pi(s) \; \big| \; s \sim \rho.
\end{equation*}
The returns for a cost, $Z_{C_i}^\pi$, $Y_{C_i}^\pi$, and $G_{C_i}^\pi$, are defined by replacing the reward with the $i$th cost.
Then, a safe RL problem can be defined as maximizing the reward return while keeping the cost returns below predefined thresholds to ensure safety. 

\subsection{Spectral Risk Measures}

In order to reduce the likelihood of encountering the worst-case scenario, it is required to incorporate risk measures into constraints.
Conditional value at risk (CVaR) is a widely used risk measure in finance \citep{rockafellar2000optimization}, and it has been further developed into a spectral risk measure in \citep{acerbi2002spectral}, which provides a more comprehensive framework.
A spectral risk measure with spectrum $\sigma$ is defined as:
\begin{equation*}
\mathcal{R}_\sigma(X) := \int_0^1 F_X^{-1}(u) \sigma(u) du,
\end{equation*}
where $\sigma$ is an increasing function, $\sigma \geq 0$, and $\int_0^1 \sigma(u)du = 1$.
By appropriately defining the function $\sigma$, various risk measures can be established, and examples are as follows:
\begin{equation}
\label{eq: example of spectral risk}
\mathrm{CVaR}_\alpha: \sigma(u) = \mathbf{1}_{u\geq\alpha}/(1-\alpha), \quad \mathrm{Pow}_\alpha: \sigma(u) = u^{\alpha/(1 - \alpha)}/(1 - \alpha),
\end{equation}
where $\alpha \in [0, 1)$ represents a risk level, and the measures become risk neutral when $\alpha=0$.
Furthermore, it is known that the spectral risk measure has the following dual form \citep{bauerle2021minimizing}:
\begin{equation}
\label{eq: spectral risk measure}
\mathcal{R}_\sigma(X) = \inf_{g} \mathbb{E}[g(X)] + \int_0^1 g^*(\sigma(u)) du,
\end{equation}
where $g:\mathbb{R} \mapsto \mathbb{R}$ is an increasing convex function, and $g^*(y):=\sup_x xy -g(x)$ is the convex conjugate function of $g$.
For example, $\mathrm{CVaR}$ is expressed as $\mathrm{CVaR}_\alpha(X) = \inf_\beta \mathbb{E}[\frac{1}{\alpha}(X-\beta)_+] + \beta$, where $g(x) = (x-\beta)_+/(1-\alpha)$, and the integral part of the conjugate function becomes $\beta$.
We define the following sub-risk measure corresponding to the function $g$ as follows:
\begin{equation}
\label{eq: sub-risk measure}
\mathcal{R}_\sigma^g(X):= \mathbb{E}[g(X)] + \int_0^1 g^*(\sigma(u)) du.
\end{equation}
Then, $\mathcal{R}_\sigma(X) = \inf_g \mathcal{R}^g_\sigma(X)$.
Note that the integral part is independent of $X$ and only serves as a constant value of risk.
As a result, the sub-risk measure can be expressed using the expectation, so it provides computational advantages over the original risk measure in many operations.
We will utilize this advantage to show optimality of the proposed method.

\subsection{Augmented State Spaces}

As mentioned by \citet{zhang2024cvar}, an risk-constrained RL problem is non-Markovian, so a history-conditioned policy is required to solve the problem.
Also, \citet{bastani2022regret} showed that an optimal history-conditioned policy can be expressed as a Markov policy defined in a newly augmented state space, which incorporates the discounted sum of costs as part of the state.
To follow these results, we define an augmented state as follows:
\begin{equation}
\label{eq: augmented state}
\begin{aligned}
\bar{s}_t := (s_t, \{e_{i,t}\}_{i=1}^N, b_t), \; \text{where} \; e_{i, 0} = 0, \; e_{i, t+1} = (C_i(s_{t}, a_t, s_{t+1}) + e_{i,t})/\gamma, \; b_t = \gamma^t.
\end{aligned}
\end{equation}
Then, the CMDP is modified as follows:
\begin{equation*}
\bar{s}_{t+1} = (s_{t+1}, \{(c_{i,t} + e_{i,t})/\gamma\}_{i=1}^N, \gamma b_t), \; \text{where} \; s_{t+1} \sim P(\cdot|s_t, a_t), \; c_{i,t} = C_i(s_t, a_t, s_{t+1}),
\end{equation*}
and the policy $\pi(\cdot|\bar{s})$ is defined on the augmented state space.

\section{Proposed Method}

The risk measure-constrained RL (RCRL) problem is defined as follows:
\begin{equation}
\label{eq: main problem}
\max_\pi \mathbb{E}[G_R^\pi] \;\; \mathbf{s.t.} \; \mathcal{R}_{\sigma_i}(G_{C_i}^\pi) \leq d_i \; \forall i \in \{1, 2, ..., N\},
\end{equation}
where $d_i$ is the threshold of the $i$th constraint.
Due to the nonlinearity of the risk measure, achieving an optimal policy through policy gradient techniques can be challenging.
To address this issue, we propose solving the RCRL problem by decomposing it into two separate problems.
Using a feasibility function $\mathcal{F}$,\footnote{$\mathcal{F}(x):= 0$ if $x\leq0$ else $\infty$.} the RCRL problem can be rewritten as follows:
\begin{equation}
\label{eq: dual form of main problem}
\begin{aligned}
\max_\pi \mathbb{E}[G_R^\pi] - \sum\nolimits_{i=1}^N \mathcal{F}(\mathcal{R}_{\sigma_i}(G_{C_i}^\pi) & - d_i) = \max_\pi \mathbb{E}[G_R^\pi] - \sum\nolimits_{i=1}^N \inf_{g_i} \mathcal{F}(\mathcal{R}^{g_i}_{\sigma_i}(G_{C_i}^\pi) - d_i) \\
&\quad = \underbrace{\sup_{\{g_i\}_{i=1}^N} \underbrace{\max_\pi \mathbb{E}[G_R^\pi] - \sum\nolimits_{i=1}^N \mathcal{F}(\mathcal{R}^{g_i}_{\sigma_i}(G_{C_i}^\pi) - d_i)}_{\textbf{(a)}}}_{\textbf{(b)}},
\end{aligned}
\end{equation}
where \textbf{(a)} is the inner problem, \textbf{(b)} is the outer problem, and $\{g_i\}_{i=1}^N$ is denoted as a dual variable.
It is known that if the cost functions are bounded and there exists a policy that satisfies the constraints of (\ref{eq: main problem}), there exists an optimal dual variable \citep{bauerle2021minimizing}, which transforms the supremum into the maximum in (\ref{eq: dual form of main problem}).
Then, the RCRL problem can be solved by \textbf{1)} finding optimal policies for various dual variables and \textbf{2)} selecting the dual variable corresponding to the policy that achieves the maximum expected reward return while satisfying the constraints.

The inner problem is then a safe RL problem with sub-risk measure constraints.
However, although the sub-risk measure is expressed using the expectation as in (\ref{eq: sub-risk measure}), the function $g$ causes nonlinearity, which makes difficult to develop a policy gradient method.
To address this issue, we propose novel risk value functions that show linearity in the performance difference between two policies.
Through these risk value functions, we propose a policy update rule with convergence guarantees for the inner problem in Section \ref{sec: policy update rule}.

The search space of the outer problem is a function space, rendering it impractical.
To address this challenge, we appropriately parameterize the function $g_i$ using a parameter $\beta_i \in B \subset \mathbb{R}^{M-1}$ in Section \ref{sec: discretization}.
Then, we can solve the outer problem through a brute-force approach by searching the parameter space.
However, it is computationally prohibitive, requiring exponential time regarding to the number of constraints and the dimension of the parameter space.
Consequently, it is necessary to develop an efficient method for solving the outer problem.
Given the difficulty of directly identifying an optimal $\beta^*$, we instead model a distribution $\xi$ to estimate the probability that a given $\beta$ is optimal.
We then propose a method that ensures convergence to an optimal distribution $\xi^*$, which deterministically samples an optimal $\beta^*$, in Section \ref{sec: outer problem solver}.

By integrating the proposed methods to solve the inner and outer problems, we finally introduce an RCRL algorithm called \textit{spectral-risk-constrained policy optimization (SRCPO)}.
The pseudo-code for SRCPO is presented in Algorithm \ref{algo: pseudo code}.
In the following sections, we will describe how to solve the inner problem and the outer problem in detail.

\begin{algorithm}[t]
\caption{Pseudo-Code of Spectral-Risk-Constrained Policy Optimization (SRCPO)}
\label{algo: pseudo code}
\small
\begin{algorithmic}
\STATE {\bfseries Input:} Spectrum function $\sigma_i$ for $i \in \{1, ..., N\}$.
\STATE Parameterize $g_i$ of $\sigma_i$ defined in (\ref{eq: sub-risk measure}) using $\beta_i \in B \subset \mathbb{R}^{M-1}$. \quad \textcolor{blue}{// Parameterization, Section \ref{sec: discretization}.}
\FOR{$t$ $=1$ {\bfseries to} $T$}
    \FOR{$\boldsymbol{\beta}=\{\beta_1, ..., \beta_N\} \in B^N$}
        \STATE Update $\pi_{\boldsymbol{\beta}, t}$ using the proposed policy gradient method. \quad \textcolor{blue}{// Inner problem, Section \ref{sec: policy update rule}.}
    \ENDFOR
    \STATE Calculate the target distribution of $\xi$ using $\pi_{\boldsymbol{\beta}, t}$, and update $\xi_t$. \quad \textcolor{blue}{// Outer problem, Section \ref{sec: outer problem solver}.}
\ENDFOR
\STATE {\bfseries Output:} $\pi_{\boldsymbol{\beta}, T}$, where $\boldsymbol{\beta} \sim \xi_T$.
\end{algorithmic}
\end{algorithm}

\section{Inner Problem: Safe RL with Sub-Risk Measure}

In this section, we propose a policy gradient method to solve the inner problem, defined as:
\begin{equation}
\label{eq: sub-problem}
\max_\pi \mathbb{E}[G_R^\pi] \;\; \mathbf{s.t.} \; \mathcal{R}^{g_i}_{\sigma_i}(G_{C_i}^\pi) \leq d_i \; \forall i,
\end{equation}
where constraints consist of the sub-risk measures, defined in (\ref{eq: sub-risk measure}).
In the remainder of this section, we first introduce risk value functions to calculate the policy gradient of the sub-risk measures, propose a policy update rule, and finally demonstrate its convergence to an optimum in tabular settings.

\subsection{Risk Value Functions}

In order to derive the policy gradient of the sub-risk measure, we first define risk value functions.
To this end, using the fact that $e_{i,t} = \sum_{j=0}^{t-1}\gamma^j c_{i,j}/\gamma^t$ and $b_t = \gamma^t$, where $\{e_{i,t}\}_{i=1}^N$ and $b_t$ is defined in the augmented state $\bar{s}_t$, we expand the sub-risk measure of the $i$th cost as follows: 
\begin{equation*}
\begin{aligned}
&\mathcal{R}_{\sigma_i}^{g_i}(G_{C_i}^\pi) - \int_0^1 {g_i}^*(\sigma_i(u))du = \int_{-\infty}^\infty g_i(z) f_{G_{C_i}^\pi}(z) dz =\mathbb{E}_{\bar{s}_0 \sim \rho} \left[ \int_{-\infty}^\infty g_i(z) f_{Y_{C_i}^\pi(\bar{s}_0)}(z) dz \right] \\
&=\underset{\rho, \pi, P}{\mathbb{E}} \left[ \int_{-\infty}^\infty g_i(z) f_{\sum_{j=0}^{t-1} \gamma^j c_{i,j} + \gamma^t Y_{C_i}^\pi(\bar{s}_t)}(z) dz \right] =\underset{\rho, \pi, P}{\mathbb{E}} \left[ \int_{-\infty}^\infty g_i(z) f_{b_t(e_{i,t} +Y_{C_i}^\pi(\bar{s}_t))}(z) dz \right] \; \forall t, \\
\end{aligned}
\end{equation*}
where $f_X$ is the probability density function (pdf) of a random variable $X$.
To capture the value of the sub-risk measure, we can define risk value functions as follows:
\begin{equation*}
V_{i, g}^\pi(\bar{s}) := \int_{-\infty}^{\infty} g(z) f_{b(e_{i}+Y_{C_i}^\pi(\bar{s}))}(z) dz, \; Q_{i, g}^\pi(\bar{s}, a) := \int_{-\infty}^{\infty} g(z)f_{b(e_{i} + Z_{C_i}^\pi(\bar{s}, a))}(z) dz.
\end{equation*}
Using the convexity of $g$, we present the following lemma on the bounds of the risk value functions:
\begin{restatable}{lemma}{RiskValueRange}
\label{lemma: range of risk value}
Assuming that a function $g$ is differentiable, $V^\pi_{i,g}(\bar{s})$ and $Q^\pi_{i,g}(\bar{s},a)$ are bounded by $[g(e_i b), g(e_i b + b C_\mathrm{max}/(1-\gamma))]$, and
$|Q^\pi_{i,g}(\bar{s},a) - V^\pi_{i,g}(\bar{s})|  \leq b C$, where $C=\frac{C_\mathrm{max}}{1-\gamma}g'(\frac{C_\mathrm{max}}{1-\gamma})$ and $\bar{s} = (s,\{e_i\}_{i=1}^N,b)$.
\end{restatable}
The proof is provided in Appendix \ref{sec: proof}.
Lemma \ref{lemma: range of risk value} means that the value of $Q^\pi_{i,g}(\bar{s},a) - V^\pi_{i,g}(\bar{s})$ is scaled by $b$.
To eliminate this scale, we define a risk advantage function as follows:
\begin{equation*}
A_{i,g}^\pi(\bar{s}, a) := (Q_{i,g}^\pi(\bar{s}, a) - V_{i,g}^\pi(\bar{s}))/b.
\end{equation*}
Now, we introduce a theorem on the difference in the sub-risk measure between two policies.
\begin{restatable}{theorem}{RMDifference}
\label{thm: risk difference}
Given two policies, $\pi$ and $\pi'$, the difference in the sub-risk measure is:
\begin{equation}
\label{eq: performance difference}
\mathcal{R}_{\sigma_i}^{g_i}(G_{C_i}^{\pi'}) - \mathcal{R}_{\sigma_i}^{g_i}(G_{C_i}^{\pi}) = \mathbb{E}_{d_\rho^{\pi'}, \pi'} \left[A_{i, g_i}^\pi (\bar{s}, a)\right]/(1-\gamma). 
\end{equation}
\end{restatable}
The proof is provided in Appendix \ref{sec: proof}.
Since Theorem \ref{thm: risk difference} also holds for the optimal policy, it is beneficial for showing optimality, as done in policy gradient algorithms of traditional RL \citep{agarwal2021theory}.
Additionally, to calculate the advantage function, we use a distributional critic instead of directly estimating the risk value functions.
This process is described in detail in Section \ref{sec: practical implementation}.

\subsection{Policy Update Rule}
\label{sec: policy update rule}

In this section, we derive the policy gradient of the sub-risk measure and introduce a policy update rule to solve the inner problem.
Before that, we first parameterize the policy as $\pi_\theta$, where $\theta \in \Theta$, and denote the objective function as $\mathbb{E}[G_R^{\pi_\theta}]=J_R(\pi) = J_R(\theta)$ and the $i$th constraint function as $\mathcal{R}_{\sigma_i}^{g_i}(G_{C_i}^{\pi_\theta}) = J_{C_i}(\pi) = J_{C_i}(\theta)$.
Then, using (\ref{eq: performance difference}) and the fact that $\mathbb{E}_{a\sim\pi(\cdot|\bar{s})}[A_{i,g}^\pi(\bar{s},a)] = 0$, the policy gradient of the $i$th constraint function is derived as follows:
\begin{equation}
\label{eq: policy gradient of sub risk measure}
\nabla_\theta J_{C_i}(\theta) = \mathbb{E}_{d_\rho^{\pi_\theta}, \pi_\theta}[A_{i, g_i}^{\pi_\theta}(\bar{s}, a) \nabla_\theta \log(\pi_\theta(a|\bar{s}))]/(1-\gamma).
\end{equation}
Now, we propose a policy update rule as follows:
\begin{equation*}
\theta_{t+1} = \begin{cases}
\theta_t + \alpha_t F^{\dagger}(\theta_t) \nabla_{\theta}\left(J_R(\theta) - \alpha_t \sum_{i=1}^N \lambda_{t,i}J_{C_i}(\theta)\right)\big|_{\theta=\theta_t} & \text{if constraints are satisfied}, \\
\theta_t + \alpha_t F^{\dagger}(\theta_t) \nabla_{\theta}\left( \alpha_t \nu_t J_R(\theta) - \sum_{i=1}^N \lambda_{t,i}J_{C_i}(\theta)\right)\big|_{\theta=\theta_t} & \text{otherwise}, \\
\end{cases}
\end{equation*}
where $\alpha_t$ is a learning rate, $F(\theta)$ is Fisher information matrix, $A^\dagger$ is Moore-Penrose inverse matrix of $A$, $\{\lambda_{t,1}, ..., \lambda_{t, N}, \nu_t\} \subset [0, \lambda_\mathrm{max}]$ are weights, and $\lambda_{t,i}=0$ for $i \in \{i|J_{C_i}(\theta) \leq d_i\}$ and $\sum_i\lambda_{t,i}=1$ when the constraints are not satisfied.
There are various possible strategies for determining $\lambda_{t, i}$ and $\nu_t$, which makes the proposed method generalize many primal approach-based safe RL algorithms \citep{achiam2017constrained, xu2021crpo, kim2023sdac}.
Several options for determining $\lambda_{t,i}$ and $\nu_t$, as well as our strategy, are described in Appendix \ref{sec: lambda strategy}.

\subsection{Convergence Analysis}

We analyze the convergence of the proposed policy update rule in a tabular setting where both the augmented state space and action space are finite.
To demonstrate convergence in this setting, we use the softmax policy parameterization \citep{agarwal2021theory} as follows:
\begin{equation*}
\pi_\theta(a|\bar{s}) := \exp{\theta(\bar{s}, a)} \big/ \left( \sum\nolimits_{a'}\exp{\theta(\bar{s}, a')} \right) \; \forall (\bar{s}, a) \in \bar{S} \times A.
\end{equation*}
Then, we can show that the policy converges to an optimal policy if updated by the proposed method.
\begin{restatable}{theorem}{convergence}
\label{thm: convergence}
Assume that there exists a feasible policy $\pi_f$ such that $J_{C_i}(\pi_f) \leq d_i - \eta$ for $\forall i$, where $\eta > 0$.
Then, if a policy is updated by the proposed update rule with a learning rate $\alpha_t$ that follows the Robbins-Monro condition \citep{robbins1951stochastic}, it will converge to an optimal policy of the inner problem.
\end{restatable}
The proof is provided in Appendix \ref{sec: proof of convergence}, and the assumption of the existence of a feasible policy is commonly used in convergence analysis in safe RL \citep{ding2020natural, bai2022achieving, kim2023sdac}.
Through this result, we can also demonstrate convergence of existing primal approach-based safe RL methods, such as CPO \citep{achiam2017constrained}, PCPO \citep{yang2020pcpo}, and P3O \citep{zhang2022penalized},\footnote{Note that these safe RL algorithms are designed to handle expectation-based constraints, so they are not suitable for RCRL.} by identifying proper $\lambda_{t,i}$ and $\nu_t$ strategies for these methods.

\section{Outer Problem: Dual Optimization for Spectral Risk Measure}

\begin{wrapfigure}{r}{0.4\linewidth}
\vspace{-20pt}
\centering
\includegraphics[width=0.8\linewidth]{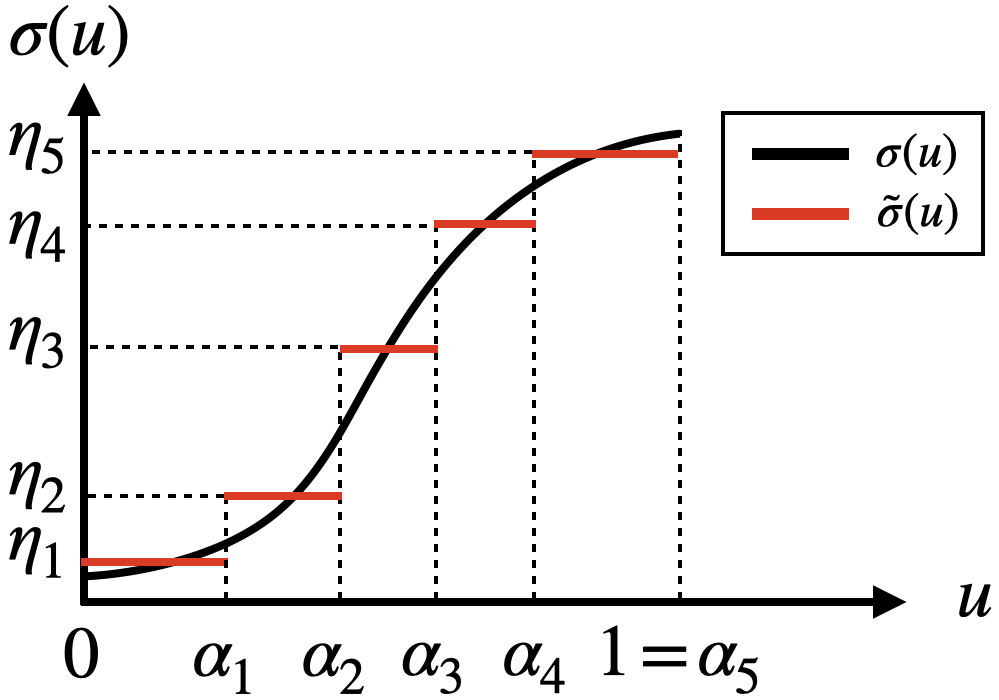}
\vspace{-5pt}
\caption{\small Discretization of spectrum.}
\label{fig: spectrum approx}
\vspace{-30pt}
\end{wrapfigure}

In this section, we propose a method for solving the outer problem that can be performed concurrently with the policy update of the inner problem.
To this end, since the domain of the outer problem is a function space, we first introduce a procedure to parameterize the functions, allowing the problem to be practically solved.
Subsequently, we propose a method to find an optimal solution for the parameterized outer problem.

\subsection{Parameterization}
\label{sec: discretization}

Searching the function space of $g$ is challenging. 
To address this issue, we discretize the spectrum function as illustrated in Figure \ref{fig: spectrum approx} since the discretization allows $g$ to be expressed using a finite number of parameters.
The discretized spectrum is then formulated as:
\begin{equation}
\label{eq: discretized spectrum}
\begin{aligned}
\sigma(u) \approx \tilde{\sigma}(u) := \eta_1 + \sum\nolimits_{i=1}^{M-1} (\eta_{i+1} - \eta_{i}) \mathbf{1}_\mathrm{u \geq \alpha_i},
\end{aligned}
\end{equation}
where $0 \leq \eta_i \leq \eta_{i+1}$, $0 \leq \alpha_{i} \leq \alpha_{i+1} \leq 1$, and $M$ is the number of discretizations.
The proper values of $\eta_i$ and $\alpha_i$ can be achieved by minimizing the norm of the function space, expressed as:
\begin{equation}
\label{eq: approx problem}
\min_{\{\eta_i\}_{i=1}^M, \{\alpha_i\}_{i=1}^{M-1}} \int_0^1 |\sigma(u) - \tilde{\sigma}(u)| du \quad \mathbf{s.t.} \; \int_0^1 \tilde{\sigma}(u) du = 1.
\end{equation}
Now, we can show that the difference between the original risk measure and its discretized version is bounded by the inverse of the number of discretizations.
\begin{restatable}[Approximation Error]{lemma}{DiffRiskMeasure}
\label{lemma: difference of discretized spectrum}
The difference between the original risk measure and its discretized version is:
\begin{equation*}
|\mathcal{R}_\sigma(X) - \mathcal{R}_{\tilde{\sigma}}(X)| \leq C_\mathrm{max}\sigma(1)/((1-\gamma)M).
\end{equation*} 
\end{restatable}
The proof is provided in Appendix \ref{sec: proof}.
Lemma \ref{lemma: difference of discretized spectrum} means that as the number of discretizations increases, the risk measure converges to the original one.
Note that the discretized version of the risk measure is also a spectral risk measure, indicating that the discretization process projects the original risk measure into a specific class of spectral risk measures.
Now, we introduce a parameterization method as detailed in the following theorem:
\begin{restatable}{theorem}{Parameterization}
\label{thm: parameterization}
Let us parameterize the function $g$ using a parameter $\beta \in B \subset \mathbb{R}^{M-1}$ as:
\begin{equation*}
g_{\beta}(x) := \eta_1 x + \sum\nolimits_{i=1}^{M-1} (\eta_{i+1} - \eta_{i})(x - \beta[i])_+,
\end{equation*}
where $\beta[i] \leq \beta[i+1]$ for $i \in \{1, .., M-2\}$.
Then, the following is satisfied:
\begin{equation*}
\mathcal{R}_{\tilde{\sigma}}(X) = \inf\nolimits_\beta \mathcal{R}^\beta_{\tilde{\sigma}}(X), \; \text{where} \; \mathcal{R}^\beta_{\tilde{\sigma}}(X) := \mathbb{E}[g_\beta(X)] + \int_0^1 g_\beta^*(\tilde{\sigma}(u))du.
\end{equation*}
\end{restatable}
According to Remark 2.7 in \citep{bauerle2021minimizing}, the infimum function in (\ref{eq: spectral risk measure}) exists and can be expressed as an integral of the spectrum function if the reward and cost functions are bounded. 
By using this fact, Theorem \ref{thm: parameterization} can be proved, and details are provided in Appendix \ref{sec: proof}.
Through this parameterization, the RCRL problem is now expressed as follows:
\begin{equation}
\label{eq: approximated RCRL}
\sup_{\{\beta_i\}_{i=1}^N} \max_\pi \mathbb{E}[G_R^\pi] - \sum\nolimits_{i=1}^N \mathcal{F}(\mathcal{R}^{\beta_i}_{\tilde{\sigma}_i}(G_{C_i}^\pi) - d_i),
\end{equation}
where $\beta_i \in B \subset \mathbb{R}^{M-1}$ is a parameter of the $i$th constraint.
In the remainder of the paper, we denote the constraint function $\mathcal{R}^{\beta_i}_{\tilde{\sigma}_i}(G_{C_i}^\pi)$ as $J_{C_i}(\pi; 
\boldsymbol{\beta})$, where $\boldsymbol{\beta} = \{\beta_i\}_{i=1}^N$, and the policy for $\boldsymbol{\beta}$ as $\pi_{\boldsymbol{\beta}}$.


\subsection{Optimization}
\label{sec: outer problem solver}


To obtain the supremum of $\boldsymbol{\beta}$ in (\ref{eq: approximated RCRL}), a brute-force search can be used, but it demands exponential time relative to the dimension of $\boldsymbol{\beta}$. 
Alternatively, $\boldsymbol{\beta}$ can be directly optimized via gradient descent; however, due to the difficulty in confirming the convexity of the outer problem, optimal convergence cannot be assured. 
To resolve this issue, we instead propose to find a distribution on $\boldsymbol{\beta}$, called a \emph{sampler} $\xi \in \mathcal{P}(B)^N$, that outputs the likelihood that a given $\boldsymbol{\beta}$ is optimal.

For detailed descriptions of the sampler $\xi$, the probability of sampling $\boldsymbol{\beta}$ is expressed as $\xi(\boldsymbol{\beta}) = \prod_{i=1}^N \xi[i](\beta_i)$, where $\xi[i]$ is the $i$th element of $\xi$, and $\beta_i$ is $\boldsymbol{\beta}[i]$.
The sampling process is denoted by $\boldsymbol{\beta}\sim\xi$, where each component is sampled according to $\beta_i \sim \xi[i]$.
Implementation of this sampling process is similar to the stick-breaking process \citep{sethuraman1994constructive} due to the condition $\beta_i[j] \leq \beta_i[j+1]$ for $j \in \{1, ..., M-2\}$.
Initially, $\beta_i[1]$ is sampled within $[0, C_\mathrm{max}/(1-\gamma)]$, and subsequent values $\beta_i[j+1]$ are sampled within $[\beta_i[j], C_\mathrm{max}/(1-\gamma)]$.
Then, our target is to find the following optimal distribution:
\begin{equation}
\label{eq: optimal distribution}
\xi^*(\boldsymbol{\beta}) \begin{cases}
\geq 0 \quad \text{if} \; \boldsymbol{\beta} \in \{\boldsymbol{\beta}|J_R(\pi_{\boldsymbol{\beta}}^*) = J_R(\pi_{\boldsymbol{\beta}^*}^*)\}, \\
= 0 \quad \text{otherwise,}
\end{cases}
\end{equation}
where $\boldsymbol{\beta}^*$ is an optimal solution of the outer problem (\ref{eq: approximated RCRL}).
Once we obtain the optimal distribution, $\boldsymbol{\beta}^*$ can be achieved by sampling from $\xi^*$.

In order to obtain the optimal distribution, we propose a novel loss function and update rule.
Before that, let us parameterize the sampler as $\xi_\phi$, where $\phi \in \Phi$, and define the following function:
\begin{equation*}
J(\pi ; \boldsymbol{\beta}):=J_R(\pi) - K \sum\nolimits_i (J_{C_i}(\pi ; \boldsymbol{\beta}) - d_i)_+.
\end{equation*} 
It is known that for sufficiently large $K > 0$, the optimal policy of the inner problem, $\pi_{\boldsymbol{\beta}}^*$, is also the solution of $\max_{\pi} J(\pi ;\boldsymbol{\beta})$, which means $J_R(\pi_{\boldsymbol{\beta}}^*) = \max_\pi J(\pi; \boldsymbol{\beta})$ \citep{zhang2022penalized}.
Using this fact, we build a target distribution for the sampler as: $\bar{\xi}_t(\boldsymbol{\beta}) \propto \exp(J(\pi_{\boldsymbol{\beta}, t}; \boldsymbol{\beta}))$, where $\pi_{\boldsymbol{\beta}, t}$ is the policy for $\boldsymbol{\beta}$ at time-step $t$.
Then, we define a loss function using the cross-entropy ($H$) as follows:
\begin{equation*}
\begin{aligned}
&\mathcal{L}_t(\phi) := H(\xi_\phi, \bar{\xi}_t) = -\mathbb{E}_{\boldsymbol{\beta} \sim \xi_\phi}[\log \bar{\xi}_t(\boldsymbol{\beta})] = -\mathbb{E}_{\boldsymbol{\beta} \sim \xi_\phi}[J(\pi_{\boldsymbol{\beta}, t}; \boldsymbol{\beta})], \\
&\nabla_\phi \mathcal{L}_t(\phi) = -\mathbb{E}_{\boldsymbol{\beta} \sim \xi_\phi}[\nabla_\phi \log(\xi_\phi(\boldsymbol{\beta})) J(\pi_{\boldsymbol{\beta}, t};\boldsymbol{\beta})].
\end{aligned}
\end{equation*}
Finally, we present an update rule along with its convergence property in the following theorem.
\begin{restatable}{theorem}{OuterConverge}
Let us assume that the space of $\boldsymbol{\beta}$ is finite and update the sampler according to the following equation:
\begin{equation}
\label{eq: sampler update}
\phi_{t+1} = \phi_t - \alpha F^\dagger(\phi_t)\nabla_\phi \mathcal{L}_t(\phi)|_{\phi=\phi_t},
\end{equation}
where $\alpha$ is a learning rate, and $F^\dagger(\phi)$ is the pseudo-inverse of Fisher information matrix of $\xi_\phi$.
Then, under a softmax parameterization, the sampler converges to an optimal distribution defined in (\ref{eq: optimal distribution}).
\end{restatable}
The proof is provided in Appendix \ref{sec: proof}.
As the loss function consists of the current policy $\pi_{\boldsymbol{\beta}, t}$, the sampler can be updated simultaneously with the policy update.

\section{Practical Implementation}
\label{sec: practical implementation}

In order to calculate the policy gradient defined in (\ref{eq: policy gradient of sub risk measure}), it is required to estimate the risk value functions.
Instead of modeling them directly, we use quantile distributional critics \citep{dabney2018quantile} $Z_{C_i, \psi}(\bar{s},a; \boldsymbol{\beta}):\bar{S}\times A \times B^N \mapsto \mathbb{R}^L$, which approximate the pdf of $Z_{C_i}^{\pi_{\boldsymbol{\beta}}}(\bar{s},a)$ using $L$ Dirac delta functions and are parameterized by $\psi \in \Psi$. 
Then, the risk value function can be approximated as:
\begin{equation*}
Q_{i, g}^{\pi_{\boldsymbol{\beta}}}(\bar{s}, a) \approx \sum\nolimits_{l=1}^L g(b e_i + b Z_{C_i, \psi}(\bar{s}, a; \boldsymbol{\beta})[l])/L,
\end{equation*}
and $V_{i, g}^{\pi_{\boldsymbol{\beta}}}(\bar{s})$ can be achieved from $\mathbb{E}_{a\sim \pi_{\boldsymbol{\beta}}(\cdot|\bar{s})}[Q_{i, g}^{\pi_{\boldsymbol{\beta}}}(\bar{s}, a)]$.
The distributional critics are trained to minimize the quantile regression loss \citep{dabney2018quantile} and details are referred to Appendix \ref{sec: implementation detail}.
Additionally, to implement $\xi_\phi$, we use a truncated normal distribution $\mathcal{N}(\mu, \sigma^2, a, b)$ \citep{burkardt2014truncated}, where $[a, b]$ is the sampling range.
Then, the sampling process of $\xi_\phi$ can be implemented as follows:
\begin{equation*}
\begin{aligned}
\beta_i[j] = \sum\nolimits_{k=1}^j \Delta\beta_i[k], \; \text{where}\; \Delta\beta_i[j] \sim \mathcal{N}(\mu_{i,\phi}[j], \sigma^2_{i,\phi}[j], 0, C_\mathrm{max}/(1-\gamma)) \; \forall i \; \text{and} \; \forall j.
\end{aligned}
\end{equation*}
Additional details of the practical implementation are provided in Appendix \ref{sec: implementation detail}.

\section{Experiments and Results}

\textbf{Tasks.}
The experiments are conducted on the Safety Gymnasium tasks \citep{ji2023safetygym} with a single constraint and the legged robot locomotion tasks \citep{kim2023sdac} with multiple constraints. 
In the Safety Gymnasium, two robots—point and car—are used to perform two tasks: a goal task, which involves controlling the robot to reach a target location, and a button task, which involves controlling the robot to press a designated button. 
In these tasks, a cost is incurred when the robot collides with an obstacle.
In the legged robot locomotion tasks, bipedal and quadrupedal robots are controlled to track a target velocity while satisfying three constraints related to body balance, body height, and foot contact timing.
For more details, please refer to Appendix \ref{sec: experimental details}.

\textbf{Baselines.}
Many RCRL algorithms use CVaR to define risk constraints \citep{zhang2024cvar, kim2023sdac, ying2022cppo}. 
Accordingly, the proposed method is evaluated and compared with other algorithms under the CVaR constraints with $\alpha = 0.75$.
Experiments on other risk measures are performed in Section \ref{sec: experiments on other risk measure}.
The baseline algorithms are categorized into three types based on their approach to estimating risk measures. 
First, CVaR-CPO \citep{zhang2024cvar} and CPPO \citep{ying2022cppo} utilize auxiliary variables to estimate CVaR. Second, WCSAC-Dist \citep{yang2023wcsac} and SDPO \citep{zhang2022sdpo} approximate the risk measure $\mathcal{R}_\sigma(G_C^\pi)$ with an expected value $\mathbb{E}_{s\sim \mathcal{D}}[\mathcal{R}_\sigma(Y_C^\pi(s))]$. 
Finally, SDAC \citep{kim2023sdac} approximates $G_C^\pi$ as a Gaussian distribution and uses the mean and standard deviation of $G_C^\pi$ to estimate CVaR.
The hyperparameters and network structure of each algorithm are detailed in Appendix \ref{sec: experimental details}.
Note that both CVaR-CPO and the proposed method, SRCPO, use the augmented state space.
However, for a fair comparison with other algorithms, the policy and critic networks of these methods are modified to operate on the original state space.

\begin{figure}[t]
    \centering
    \includegraphics[width=1\linewidth]{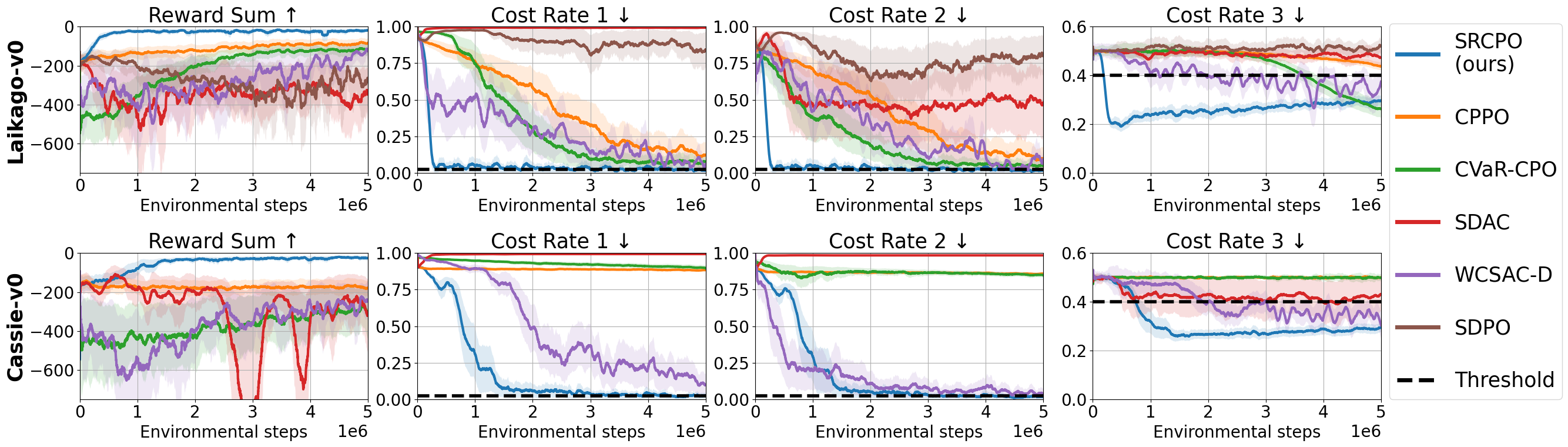}
    \caption{\small \textbf{Training curves of the legged robot locomotion tasks.}
    The upper graph shows results for the quadrupedal robot, and the lower one is for the bipedal robot. 
    The solid line in each graph represents the average of each metric, and the shaded area indicates the standard deviation scaled by $0.5$. 
    The results are obtained by training each algorithm with five random seeds.
    }
    \label{fig: legged robot results}
\end{figure}

\begin{figure}[t]
    \centering
    \includegraphics[width=1\linewidth]{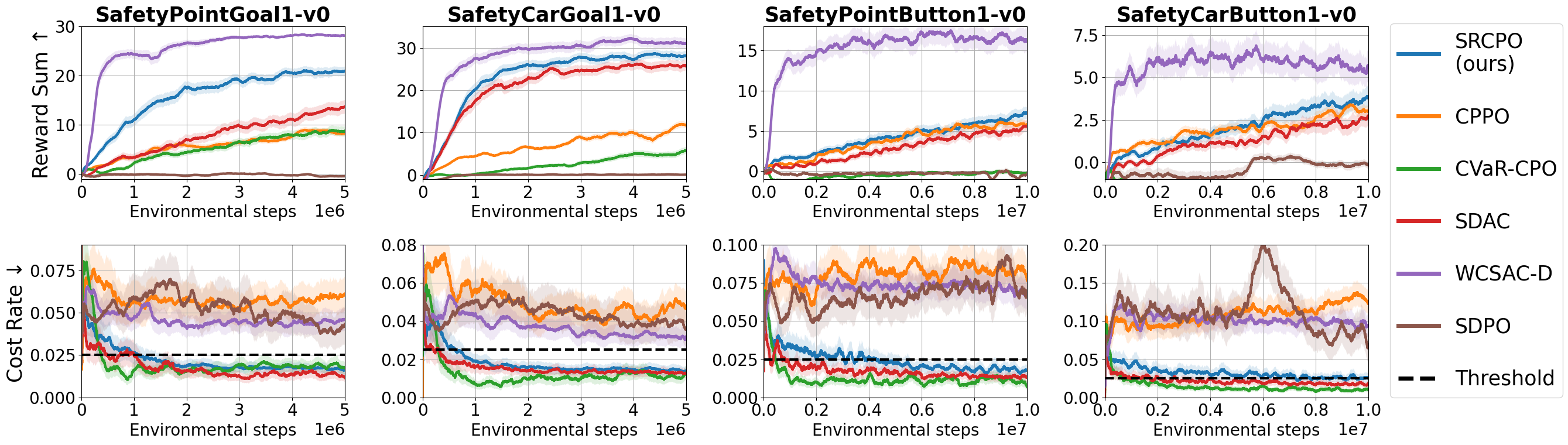}
    \caption{\small \textbf{Training curves of the Safety Gymnasium tasks.}
    The results for each task are displayed in columns, titled with the task name.
    The solid line represents the average of each metric, and the shaded area indicates the standard deviation scaled by $0.2$. 
    The results are obtained by training each algorithm with five random seeds.    
    }
    \label{fig: safety gym results}
\end{figure}

\textbf{Results.}
Figures \ref{fig: legged robot results} and \ref{fig: safety gym results} show the training curves for the locomotion tasks and the Safety Gymnasium tasks, respectively.
The reward sum in the figures refers to the sum of rewards within an episode, while the cost rate is calculated as the sum of costs divided by the episode length. 
In all tasks, the proposed method achieves the highest rewards among methods whose cost rates are below the specified thresholds.
These results are likely because only the proposed method guarantees optimality. 
Specifically in the locomotion tasks, an initial policy often struggles to stabilize the balance of the robot, resulting in high costs from the start.
Given this challenge, it is more susceptible to falling into local optima compared to other tasks, which enables the proposed method to outperform other baseline methods.
Note that WCSAC-Dist shows the highest rewards in the Safety Gymnasium tasks, but the cost rates exceed the specified thresholds. 
This issue seems to arise from the approach to estimating risk constraints. 
WCSAC-Dist estimates the constraints based on the expected risk for each state, but it is lower than the original risk measure, leading to constraint violations.

\subsection{Study on Various Risk Measures}
\label{sec: experiments on other risk measure}


\begin{figure}[t]
    \centering
    \begin{subfigure}[b]{0.34\textwidth}
        \centering
        \includegraphics[width=\textwidth]{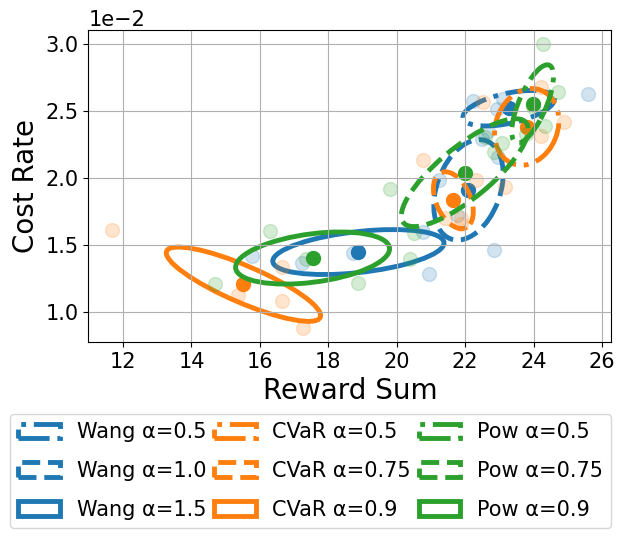}
    \end{subfigure}
    \hfill
    \begin{subfigure}[b]{0.315\textwidth}
        \centering
        \includegraphics[width=\textwidth]{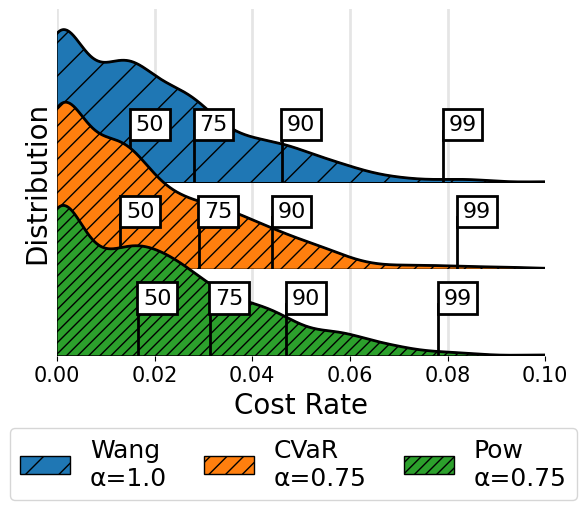}
    \end{subfigure}
    \hfill
    \begin{subfigure}[b]{0.315\textwidth}
        \centering
        \includegraphics[width=\textwidth]{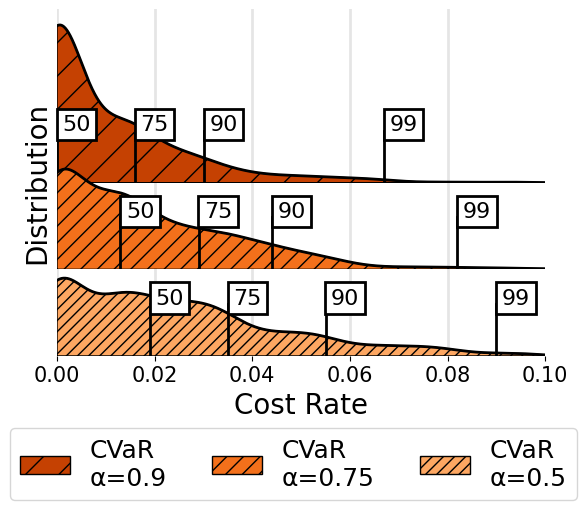}
    \end{subfigure}
    \caption{\small 
    \textbf{(Left)} A correlation graph between cost rate and reward sum for policies trained in the point goal task under various risk measure constraints. 
    The results are achieved by training policies with five random seeds for each risk measure and risk level. 
    The center and radius of each ellipse show the average and standard deviation of the results from the five seeds, respectively.
    \textbf{(Middle)} Distribution graphs of the cost rate under different risk measure constraints.
    Locations of several percentiles (from the $50$th to the $99$th) are marked on the plot.
    The risk level of each risk measure is selected to have a similar cost rate.
    After training a policy in the point goal task, cost distributions have been collected by rolling out the trained policy across 500 episodes.
    \textbf{(Right)} Distribution graphs of the cost rate with different risk levels, $\alpha$, under the CVaR constraint.
    }
    \label{fig: ablation}
    \vspace{-10pt}
\end{figure}

In this section, we analyze the results when constraints are defined using various risk measures. 
To this end, we train policies in the point goal task under constraints based on the $\mathrm{CVaR}_\alpha$ and $\mathrm{Pow}_\alpha$ risk measures defined in (\ref{eq: spectral risk measure}), as well as the Wang risk measure \citep{wang2000class}.
Although the Wang risk measure is not a spectral but a distortion risk measure, our parameterization method introduced in Section \ref{sec: discretization} enables it to be approximated as a spectral risk measure, and the visualization of this process is provided in Appendix \ref{sec: additional experimental results}.
We conduct experiments with three risk levels for each risk measure and set the constraint threshold as $0.025$.
Evaluation results are presented in Figure \ref{fig: ablation}, and training curves are provided in Appendix \ref{sec: additional experimental results}.
Figure \ref{fig: ablation} (Right) shows intuitive results that increasing the risk level effectively reduces the likelihood of incurring high costs.
Similarly, Figure \ref{fig: ablation} (Left) presents the trend across all risk measures, indicating that higher risk levels correspond to lower cost rates and decreased reward performance.
Finally, Figure \ref{fig: ablation} (Middle) exhibits the differences in how each risk measure addresses worst-case scenarios.
In the spectrum formulation defined in (\ref{eq: example of spectral risk}), $\mathrm{CVaR}$ applies a uniform penalty to the tail of the cost distribution above a specified percentile, whereas measures like $\mathrm{Wang}$ and $\mathrm{Pow}$ impose a heavier penalty at higher percentiles.
As a result, because $\mathrm{CVaR}$ imposes a relatively milder penalty on the worst-case outcomes compared to other risk measures, it shows the largest intervals between the $50$th and $99$th percentiles.

\section{Conclusions and Limitations}

In this work, we introduced a spectral-risk-constrained RL algorithm that ensures convergence and optimality in a tabular setting. Specifically, in the inner problem, we proposed a generalized policy update rule that can facilitate the development of a new safe RL algorithm with convergence guarantees. 
For the outer problem, we introduced a notion called \emph{sampler}, which enhances training efficiency by concurrently training with the inner problem.
Through experiments in continuous control tasks, we empirically demonstrated the superior performance of the proposed method and its capability to handle various risk measures.
However, convergence to an optimum is shown only in a tabular setting, so future research may focus on extending these results to linear MDPs or function approximation settings.
Furthermore, since our approach can be applied to risk-sensitive RL, future work can also implement the proposed method in this area.



\bibliographystyle{plainnat}
\bibliography{main}


\newpage
\appendix

\section{Proofs}
\label{sec: proof}

\RiskValueRange*
\begin{proof}
Since $0 \leq Y_{C_i}^\pi(\bar{s}), Z_{C_i}^\pi(\bar{s},a) \leq C_\mathrm{max}/(1-\gamma)$,
\begin{equation*}
f_{b e_i + b Y_{C_i}^\pi(\bar{s})}(z) = f_{b e_i + b Z_{C_i}^\pi(\bar{s},a)}(z) = 0,
\end{equation*}
when $z < b e_i$ or $z > be_i + bC_\mathrm{max}/(1-\gamma)$.
Using the fact that $g$ is an increasing and convex function,
\begin{equation*}
g(be_i) \leq V_{i,g}^\pi(\bar{s}) = \int_{b e_i}^{be_i + b\frac{C_\mathrm{max}}{1-\gamma}} g(z) f_{be_i + b Y_{C_i}^\pi(\bar{s})}(z) dz \leq g(be_i + bC_\mathrm{max}/(1-\gamma)).
\end{equation*}
Likewise, $Q_{i,g}^\pi(\bar{s}, a)$ also satisfies the above property.
As a result,
\begin{equation*}
V_{i,g}^\pi(\bar{s}), \; Q_{i,g}^\pi(\bar{s}, a) \in [g(be_i), g(be_i + bC_\mathrm{max}/(1-\gamma))].    
\end{equation*}
Then, the maximum of $|Q_{i,g}^\pi(\bar{s}, a) - V_{i,g}^\pi(\bar{s})|$ is $g(be_i + bC_\mathrm{max}/(1-\gamma)) - g(be_i)$.
Since $b = \gamma^t$ and $e_i = \sum_{k=0}^{t-1} \gamma^k C_i(s_k, a_k, s_{k+1})/\gamma^t$, where $t$ is the current time step,
\begin{equation*}
be_i + bC_\mathrm{max}/(1-\gamma) \leq C_\mathrm{max}/(1-\gamma).
\end{equation*}
Consequently,
\begin{equation*}
\begin{aligned}
&|Q_{i,g}^\pi(\bar{s}, a) - V_{i,g}^\pi(\bar{s})| \leq g(be_i + bC_\mathrm{max}/(1-\gamma)) - g(be_i) \\
&\quad \leq b\frac{C_\mathrm{max}}{1-\gamma} g'\left(be_i + b\frac{C_\mathrm{max}}{1-\gamma}\right) \leq b\frac{C_\mathrm{max}}{1-\gamma} g'\left(\frac{C_\mathrm{max}}{1-\gamma}\right).
\end{aligned}
\end{equation*}
\end{proof}

\RMDifference*
\begin{proof}
Since the following equation is satisfied:
\begin{equation*}
\begin{aligned}
\int_{-\infty}^\infty g(z)f_{be_i + bZ_{C_i}^\pi(\bar{s}, a)}(z)dz &= \mathbb{E}_{\bar{s}' \sim P(\cdot|\bar{s},a)} \left[ \int_{-\infty}^\infty g(z)f_{b(e_i + C_i(s,a,s')) + \gamma bY_{C_i}^\pi(\bar{s}')}(z)dz \right] \\
&= \mathbb{E}_{\bar{s}' \sim P(\cdot|\bar{s},a)} \left[ \int_{-\infty}^\infty g(z)f_{b'e_i' + b'Y_{C_i}^\pi(\bar{s}')}(z)dz \right],
\end{aligned}
\end{equation*}
we can say $Q_{i,g}^\pi(\bar{s},a)=\mathbb{E}_{\bar{s}' \sim P(\cdot|\bar{s},a)}[V_{i,g}^\pi(\bar{s}')]$.
Using this property,
\begin{equation}
\label{eq: risk diff 1}
\begin{aligned}
\mathbb{E}_{d_\rho^{\pi'}, \pi'}\left[A_{i, g_i}^\pi (\bar{s}, a)\right]/(1-\gamma) &= \mathbb{E}_{\pi'}\left[\sum_{t=0}^\infty \gamma^t A_{i, g_i}^\pi (\bar{s}_t, a_t)\right] \\
&= \mathbb{E}_{\pi'}\left[\sum_{t=0}^\infty Q_{i, g_i}^\pi (\bar{s}_t, a_t) - V_{i, g_i}^\pi (\bar{s}_t)\right] \\
&= \mathbb{E}_{\pi'}\left[\sum_{t=0}^\infty \left( V_{i, g_i}^\pi(\bar{s}_{t+1}) - V_{i, g_i}^\pi(\bar{s}_{t}) \right) \right] \\
&= \lim_{t \to \infty} \mathbb{E}_{\pi'}\left[V_{i, g_i}^\pi(\bar{s}_t)\right] - \mathbb{E}_{\bar{s}_0 \sim \rho}\left[V_{i, g_i}^\pi(\bar{s}_0)\right] \\
&= \lim_{t \to \infty} \mathbb{E}_{\pi'}\left[V_{i, g_i}^\pi(\bar{s}_t)\right] - \mathcal{R}_{\sigma_i}^{g_i}(G_{C_i}^{\pi}) + \int_0^1 {g_i}^*(\sigma_i(u))du.
\end{aligned}
\end{equation}
Using the definition of $V_{i, g_i}^\pi(\bar{s}_t)$ and the dominated convergence theorem,
\begin{equation}
\label{eq: risk diff 2}
\begin{aligned}
\lim_{t \to \infty} \mathbb{E}_{\pi'}\left[V_{i, g_i}^\pi(\bar{s}_t)\right] &= \lim_{t \to \infty} \mathbb{E}_{\pi'}\left[\int_{-\infty}^\infty g_i(z)f_{b_te_{i,t} + b_t Y_{C_i}^\pi(\bar{s}_t)}(z)dz\right] \\
&= \lim_{t \to \infty} \mathbb{E}_{\pi'}\left[\int_{-\infty}^\infty g_i(z)f_{\sum_{k=0}^{t-1} \gamma^kC_i(s_k, a_k, s_{k+1}) + \gamma^t Y_{C_i}^\pi(\bar{s}_t)}(z)dz\right] \\
&= \mathbb{E}_{\pi'}\left[\int_{-\infty}^\infty g_i(z)f_{G_{C_i}^{\pi'}}(z)dz\right] = \mathcal{R}_{\sigma_i}^{g_i}(G_{C_i}^{\pi'}) - \int_0^1 {g_i}^*(\sigma_i(u))du.
\end{aligned}
\end{equation}
By combining (\ref{eq: risk diff 1}) and (\ref{eq: risk diff 2}),
\begin{equation*}
\begin{aligned}
\mathbb{E}_{d_\rho^{\pi'}, \pi'}\left[A_{i, g_i}^\pi (\bar{s}, a)\right]/(1-\gamma) &= \lim_{t \to \infty} \mathbb{E}_{\pi'}\left[V_{i, g_i}^\pi(\bar{s}_t)\right] - \mathcal{R}_{\sigma_i}^{g_i}(G_{C_i}^{\pi}) + \int_0^1 {g_i}^*(\sigma_i(u))du \\
&= \mathcal{R}_{\sigma_i}^{g_i}(G_{C_i}^{\pi'}) - \mathcal{R}_{\sigma_i}^{g_i}(G_{C_i}^{\pi}). 
\end{aligned}
\end{equation*}
\end{proof}

\DiffRiskMeasure*
\begin{proof}
The difference is:
\begin{equation*}
|\mathcal{R}_\sigma(X) - \mathcal{R}_{\tilde{\sigma}}(X)| \leq \int_0^1 | F_X^{-1}(u)| |\sigma(u) - \tilde{\sigma}(u)| du \leq ||\sigma - \tilde{\sigma}||_1 ||F_X^{-1}||_\infty = ||\sigma - \tilde{\sigma}||_1 \frac{C_\mathrm{max}}{1-\gamma}.
\end{equation*}
The value of (\ref{eq: approx problem}) is $||\sigma - \tilde{\sigma}||_1$ and is smaller than the value of the following problem:
\begin{equation}
\label{eq: upper approx problem}
\min_{\eta_i, \alpha_i} \int_0^1 |\sigma(u) - \tilde{\sigma}(u)| du \quad \mathbf{s.t.} \; \int_0^1 \tilde{\sigma}(u) du = 1, \; \frac{\sigma(1)(i-1)}{M} \leq \eta_i \leq \frac{\sigma(1)i}{M} \; \forall i,
\end{equation}
since the search space is smaller than the original one.
As the value of (\ref{eq: upper approx problem}) is smaller than $\max_u |\sigma(u) - \tilde{\sigma}(u)| \leq \sigma(1)/M$, $||\sigma - \tilde{\sigma}||_1 \leq \sigma(1)/M$.
Consequently,
\begin{equation*}
|\mathcal{R}_\sigma(X) - \mathcal{R}_{\tilde{\sigma}}(X)|  \leq ||\sigma - \tilde{\sigma}||_1 \frac{C_\mathrm{max}}{1-\gamma} \leq \frac{C_\mathrm{max} \sigma(1)}{(1-\gamma)M}.
\end{equation*}
\end{proof}

\Parameterization*
\begin{proof}
Since $g_\beta(x)$ is an increasing convex function, $\mathcal{R}_{\tilde{\sigma}}(X) \leq \mathcal{R}_{\tilde{\sigma}}^{g_\beta}(X)=:\mathcal{R}_{\tilde{\sigma}}^\beta(X)$.
In addition, according to Remark 2.7 in \citep{acerbi2002spectral}, there exists $\tilde{g}$ such that $\tilde{g}= \arg\inf_g\mathcal{R}^g_{\tilde{\sigma}}(X)$, and its derivative is expressed as $\tilde{g}'= \tilde{\sigma}(F_X(x))$.
Using this fact, we can formulate $\tilde{g}$ by integrating its derivative as follows:
\begin{equation*}
\tilde{g}(x) = \eta_1 x + \sum_{i=1}^{M-1}(\eta_{i+1} - \eta_i)(x - F_X^{-1}(\alpha_i))_+ + C,
\end{equation*}
where $C$ is a constant.
Since for any constant $C$, $\mathcal{R}^{\tilde{g}}_{\tilde{\sigma}}(X)$ has the same value due to the integral part of $\tilde{g}^*$ in (\ref{eq: sub-risk measure}), we set $C$ as zero.
Also, we can express $\tilde{g}(x)$ as $g_{\tilde{\beta}}(x)$, where ${\tilde{\beta}}[i]$ is $F_X^{-1}(\alpha_i)$ due to the definition of $g_\beta$.
As a result, the following is satisfied:
\begin{equation*}
\begin{aligned}
\mathcal{R}_{\tilde{\sigma}}(X) = \mathcal{R}^{\tilde{g}}_{\tilde{\sigma}}(X) &= \mathcal{R}^{\tilde{\beta}}_{\tilde{\sigma}}(X) \leq \mathcal{R}_{\tilde{\sigma}}^\beta(X). \\
\Rightarrow \mathcal{R}_{\tilde{\sigma}}(X) &= \inf_\beta \mathcal{R}_{\tilde{\sigma}}^\beta(X).
\end{aligned}
\end{equation*}

\end{proof}

\OuterConverge*
\begin{proof}
Since we assume that the space of $\boldsymbol{\beta}$ is finite, the sampler can be expressed using the softmax parameterization as follows:
\begin{equation*}
\xi_\phi(\boldsymbol{\beta}) := \frac{\exp(\phi(\boldsymbol{\beta}))}{\sum_{\boldsymbol{\beta}'} \exp(\phi(\boldsymbol{\beta}'))}.
\end{equation*}
If we update the parameter of the sampler as described in (\ref{eq: sampler update}), the following is satisfied as in the natural policy gradient:
\begin{equation*}
\phi_{t+1}(\boldsymbol{\beta}) = \phi_{t}(\boldsymbol{\beta}) + \alpha J(\pi_{\boldsymbol{\beta}, t} ; \boldsymbol{\beta}) = \sum_{k=1}^t \alpha J(\pi_{\boldsymbol{\beta}, k}; \boldsymbol{\beta}).
\end{equation*}
Furthermore, due to Theorem \ref{thm: convergence}, if a policy $\pi_{\boldsymbol{\beta}, t}$ is updated by the proposed method, the policy converges to an optimal policy $\pi^*_{\boldsymbol{\beta}, t}$.
Then, if $\boldsymbol{\beta}$ is not optimal,
\begin{equation*}
\lim_{T\to\infty} \sum_{t=1}^{T} \alpha \left( J(\pi_{\boldsymbol{\beta}, t}; \boldsymbol{\beta}) - J(\pi_{\boldsymbol{\beta}^*, t}; \boldsymbol{\beta}^*) \right) = -\infty,
\end{equation*}
since $\lim_{t\to\infty} J(\pi_{\boldsymbol{\beta},t}; \boldsymbol{\beta}) - J(\pi_{\boldsymbol{\beta}^*,t}; \boldsymbol{\beta}^*) = J(\pi_{\boldsymbol{\beta}}^*; \boldsymbol{\beta}) - J(\pi_{\boldsymbol{\beta}^*}^*; \boldsymbol{\beta}^*) < 0$.
As a result,
\begin{equation*}
\begin{aligned}
&\lim_{t\to\infty}\xi_{\phi_t}(\boldsymbol{\beta}) = \lim_{t\to\infty} \exp(\phi_t(\boldsymbol{\beta}) - \phi_t(\boldsymbol{\beta}^*))/\sum_{\boldsymbol{\beta}'}\exp(\phi_t(\boldsymbol{\beta}') - \phi_t(\boldsymbol{\beta}^*)) \\
&= \lim_{t\to\infty} \exp \left( \sum_{k=1}^{t-1} \alpha \left( J(\pi_{\boldsymbol{\beta}, k}; \boldsymbol{\beta}) - J(\pi_{\boldsymbol{\beta}^*, k}; \boldsymbol{\beta}^*) \right) \right) \Big/ \sum_{\boldsymbol{\beta}'}\exp(\phi_t(\boldsymbol{\beta}') - \phi_t(\boldsymbol{\beta}^*)) = 0.
\end{aligned}
\end{equation*}
Thus, $\xi_{\phi_t}$ converges to an optimal sampler.
\end{proof}

\subsection{Proof of Theorem \ref{thm: convergence}}
\label{sec: proof of convergence}

Our derivation is based on the policy gradient works by \citet{agarwal2021theory} and \citet{xu2021crpo}.
Due to the softmax policy parameterization, the following is satisfied:
\begin{equation}
\label{eq: next poliy exression}
\begin{aligned}
\pi_{t+1}(a|\bar{s}) &= \pi_{t}(a|\bar{s}) \frac{\exp(\theta_{t+1}(\bar{s},a) - \theta_t(\bar{s},a))}{Z_t(\bar{s})},
\end{aligned}
\end{equation}
where $Z_t(\bar{s}) := \sum_{a\in A} \pi_t(a|\bar{s})\exp(\theta_{t+1}(\bar{s},a) - \theta_t(\bar{s},a))$.
By the natural policy gradient,
\begin{equation}
\label{eq: natural policy gradient update}
\theta_{t+1} = \begin{cases}
\theta_t + \alpha_t(A_R^t - \alpha_t \sum_i \lambda_{t,i} A_{i, g_i}^t)/(1-\gamma) & \text{if} \; J_{C_i}(\theta_t) \leq d_i \; \forall i,\\
\theta_t + \alpha_t(\alpha_t \nu_t A_R^t - \sum_i \lambda_{t,i} A_{i, g_i}^t)/(1-\gamma) & \text{otherwise}.
\end{cases}
\end{equation}
Before proving Theorem \ref{thm: convergence}, we first present several useful lemmas.
\begin{lemma}
\label{lemma: log z}
$\log Z_t(\bar{s})$ is non-negative.
\end{lemma}
\begin{proof}
Due to the advantage functions in (\ref{eq: natural policy gradient update}),
$\sum_{a \in A}\pi_t(a|\bar{s})(\theta_{t+1}(\bar{s}, a) - \theta_{t}(\bar{s}, a)) = 0$.
Using the fact that a logarithmic function is concave,
\begin{equation*}
\log Z_t(\bar{s}) = \log \sum_{a\in A} \pi_t(a|\bar{s}) \exp(\theta_{t+1}(\bar{s}, a) - \theta_{t}(\bar{s}, a)) \geq \sum_{a\in A} \pi_t(a|\bar{s}) (\theta_{t+1}(\bar{s}, a) - \theta_{t}(\bar{s}, a)) = 0.
\end{equation*}
As a result, $\log Z_t(\bar{s}) \geq 0$.
\end{proof}

\begin{lemma}
\label{lemma: TV inequality}
$D_\mathrm{TV}(\pi_{t+1}(\cdot|\bar{s}), \pi_{t}(\cdot|\bar{s})) \leq \max_a|\theta_{t+1}(\bar{s}, a) - \theta_{t}(\bar{s}, a)| \leq ||\theta_{t+1} - \theta_{t}||_\infty$.
\end{lemma}
\begin{proof}
Using the Pinsker's inequality,
\begin{equation*}
\begin{aligned}
&D_\mathrm{TV}(\pi_{t+1}(\cdot|\bar{s}), \pi_{t}(\cdot|\bar{s})) \leq \sqrt{\frac{1}{2} D_\mathrm{KL}(\pi_{t+1}(\cdot|\bar{s}) || \pi_{t}(\cdot|\bar{s}))} \\
&= \sqrt{\frac{1}{2} \sum_a \pi_{t+1}(a|\bar{s}) \log (\pi_{t+1}(a|\bar{s}) / \pi_{t}(a|\bar{s})} ) \\
&= \sqrt{\frac{1}{2} (\sum_a \pi_{t+1}(a|\bar{s}) (\theta_{t+1}(\bar{s}, a) - \theta_{t}(\bar{s}, a)) - \log Z_t(\bar{s}))} \\
&\overset{\text{(i)}}{\leq} \sqrt{\frac{1}{2} \sum_a \pi_{t+1}(a|\bar{s}) (\theta_{t+1}(\bar{s}, a) - \theta_{t}(\bar{s}, a))} \\
&\overset{\text{(ii)}}{=} \sqrt{\frac{1}{2}\sum_a (\pi_{t+1}(a|\bar{s}) - \pi_{t}(a|\bar{s})) (\theta_{t+1}(\bar{s}, a) - \theta_{t}(\bar{s}, a))} \\
&\leq \sqrt{\max_a|\theta_{t+1}(\bar{s}, a) - \theta_{t}(\bar{s}, a)| D_\mathrm{TV}(\pi_{t+1}(a|\bar{s}), \pi_{t}(a|\bar{s}))}, \\
\end{aligned}
\end{equation*}
where (i) follows Lemma \ref{lemma: log z}, and (ii) follows the fact that $\sum_{a \in A}\pi_t(a|\bar{s})(\theta_{t+1}(\bar{s}, a) - \theta_{t}(\bar{s}, a)) = 0$.
As a result,
\begin{equation*}
D_\mathrm{TV}(\pi_{t+1}(\cdot|\bar{s}), \pi_{t}(\cdot|\bar{s})) \leq \max_a|\theta_{t+1}(\bar{s}, a) - \theta_{t}(\bar{s}, a)| \leq ||\theta_{t+1} - \theta_{t}||_\infty.
\end{equation*}
\end{proof}

\convergence*
\begin{proof}
We divide the analysis into two cases: 1) when the constraints are satisfied and 2) when the constraints are violated.
First, let us consider the case where the constraints are satisfied.
In this case, the policy is updated as follows:
\begin{equation}
\label{eq: next policy parameter1}
\begin{aligned}
\theta_{t+1} &= \theta_t + \frac{\alpha_t}{1-\gamma} \left(A_R^t - \alpha_t \sum_i \lambda_{t,i} A_{i, g_i}^t\right). \\ 
\end{aligned}
\end{equation}
Using Theorem \ref{thm: risk difference},
\begin{equation}
\label{eq: converegence proof 11}
\begin{aligned}
&J_R(\theta_{t+1}) - J_R(\theta_t) + \alpha_t\sum_i \lambda_{t,i}(J_{C_i}(\theta_t) -  J_{C_i}(\theta_{t+1})) \\
&= \frac{1}{1 - \gamma} \sum_{\bar{s}} d_\rho^{\pi_{t+1}}(\bar{s}) \sum_a \pi_{t+1}(a|\bar{s}) (A_R^t(\bar{s}, a) - \alpha_t \sum_i \lambda_{t,i} A_{i, g_i}^t(\bar{s}, a)) \\
&= \frac{1}{1 - \gamma} \sum_{\bar{s}} d_\rho^{\pi_{t+1}}(\bar{s}) \sum_a (\pi_{t+1}(a|\bar{s}) - \pi_{t}(a|\bar{s})) (A_R^t(\bar{s}, a) - \alpha_t \sum_i \lambda_{t,i} A_{i, g_i}^t(\bar{s}, a)) \\
&\overset{\text{(i)}}{\leq} \frac{2}{1 - \gamma} \sum_{\bar{s}} d_\rho^{\pi_{t+1}}(\bar{s}) \max_a \big| A_R^t(\bar{s}, a) - \alpha_t \sum_i \lambda_{t,i} A_{i, g_i}^t(\bar{s}, a) \big| D_\mathrm{TV}(\pi_{t+1}(\cdot|\bar{s}), \pi_{t}(\cdot|\bar{s})) \\
&\overset{\text{(ii)}}{\leq} \frac{2 ||\theta_{t+1} - \theta_{t}||_\infty}{1 - \gamma} \sum_{\bar{s}} d_\rho^{\pi_{t+1}}(\bar{s}) \max_a \big| A_R^t(\bar{s}, a) - \alpha_t \sum_i \lambda_{t,i} A_{i, g_i}^t(\bar{s}, a) \big| \\
&\leq \frac{2 ||\theta_{t+1} - \theta_{t}||_\infty}{1 - \gamma} \big|\big| A_R^t - \alpha_t \sum_i \lambda_{t,i} A_{i, g_i}^t \big|\big|_\infty = \frac{2\alpha_t}{(1-\gamma)^2} \big|\big| A_R^t - \alpha_t \sum_i \lambda_{t,i} A_{i, g_i}^t \big|\big|_\infty^2 \\
&\leq 2 \alpha_t (R_\mathrm{max} + \alpha_t N\lambda_\mathrm{max}C_\mathrm{max})^2/(1-\gamma)^4.
\end{aligned}
\end{equation}
where (i) follows Hölder's inequality, and (ii) follows Lemma \ref{lemma: TV inequality}.
Using (\ref{eq: next poliy exression}) and (\ref{eq: next policy parameter1}),
\begin{equation}
\label{eq: convergence proof 12}
\begin{aligned}
&J_R(\theta_{t+1}) - J_R(\theta_t) + \alpha_t\sum_i \lambda_{t,i}(J_{C_i}(\theta_t) -  J_{C_i}(\theta_{t+1})) \\
&= \frac{1}{1-\gamma} \sum_{\bar{s}\in \bar{S}}d_\rho^{\pi_{t+1}}(\bar{s}) \sum_{a \in A}\pi_{t+1}(a|\bar{s}) (A_R^t(\bar{s}, a) - \alpha_t \sum_i \lambda_{t,i} A_{i, g_i}^t(\bar{s}, a)) \\
&= \frac{1}{\alpha_t} \sum_{\bar{s}\in \bar{S}}d_\rho^{\pi_{t+1}}(\bar{s}) \sum_{a \in A}\pi_{t+1}(a|\bar{s}) \log (\pi_{t+1}(a|\bar{s}) Z_t(\bar{s})/\pi_t(a|\bar{s})) \\
&= \frac{1}{\alpha_t} \sum_{\bar{s}\in \bar{S}}d_\rho^{\pi_{t+1}}(\bar{s}) (D_\mathrm{KL}(\pi_{t+1}(\cdot|\bar{s}) || \pi_{t}(\cdot|\bar{s})) + \log Z_t(\bar{s})) \\
& \geq \frac{1}{\alpha_t} \sum_{\bar{s}\in \bar{S}}d_\rho^{\pi_{t+1}}(\bar{s}) \log Z_t(\bar{s}) \overset{\text{(i)}}{\geq} \frac{1 - \gamma}{\alpha_t} \sum_{\bar{s}\in \bar{S}}\rho(\bar{s}) \log Z_t(\bar{s}),
\end{aligned}
\end{equation}
where (i) is from the fact that $d_\rho^\pi(\bar{s}) = (1-\gamma)\sum_t \gamma^t \mathbb{P}(\bar{s}_t=\bar{s}) \geq (1-\gamma)\rho(\bar{s})$.
By combining (\ref{eq: converegence proof 11}) and (\ref{eq: convergence proof 12}), the following is satisfied for any $\rho$:
\begin{equation}
\label{eq: convergence proof 13}
\begin{aligned}
\frac{1 - \gamma}{\alpha_t} \sum_{\bar{s}\in \bar{S}}\rho(\bar{s}) \log Z_t(\bar{s}) \leq \frac{2\alpha_t}{(1-\gamma)^4}(R_\mathrm{max} + \alpha_t N\lambda_\mathrm{max}C_\mathrm{max})^2.
\end{aligned}
\end{equation}
Then, the following is satisfied for any policy $\hat{\pi}$:
\begin{equation}
\label{eq: unviolated case inequality}
\begin{aligned}
&J_R(\hat{\pi}) - J_R(\pi_t) + \alpha_t\sum_i \lambda_{t,i}(J_{C_i}(\pi_t) -  J_{C_i}(\hat{\pi})) \\
&= \frac{1}{1-\gamma} \sum_{\bar{s}\in \bar{S}}d_\rho^{\hat{\pi}}(\bar{s}) \sum_{a \in A}\hat{\pi}(a|\bar{s}) (A_R^t(\bar{s}, a) - \alpha_t \sum_i \lambda_{t,i} A_{i, g_i}^t(\bar{s}, a)) \\
&= \frac{1}{\alpha_t} \sum_{\bar{s}\in \bar{S}}d_\rho^{\hat{\pi}}(\bar{s}) \sum_{a \in A}\hat{\pi}(a|\bar{s}) \log (\pi_{t+1}(a|\bar{s}) Z_t(\bar{s}) / \pi_t(a|\bar{s})) \\
&= \frac{1}{\alpha_t} \sum_{\bar{s}\in \bar{S}}d_\rho^{\hat{\pi}}(\bar{s}) (D_\mathrm{KL}(\hat{\pi}(\cdot|\bar{s}) || \pi_{t}(\cdot|\bar{s})) - D_\mathrm{KL}(\hat{\pi}(\cdot|\bar{s}) || \pi_{t+1}(\cdot|\bar{s})) + \log Z_t(\bar{s})) \\
& \overset{\text{(i)}}{\leq} \frac{1}{\alpha_t} \sum_{\bar{s}\in \bar{S}}d_\rho^{\hat{\pi}}(\bar{s}) (D_\mathrm{KL}(\hat{\pi}(\cdot|\bar{s}) || \pi_{t}(\cdot|\bar{s})) - D_\mathrm{KL}(\hat{\pi}(\cdot|\bar{s}) || \pi_{t+1}(\cdot|\bar{s}))) \\
&+ \frac{2\alpha_t}{(1-\gamma)^5}(R_\mathrm{max} + \alpha_t N\lambda_\mathrm{max}C_\mathrm{max})^2,
\end{aligned}
\end{equation}
where (i) follows (\ref{eq: convergence proof 13}) by substituting $\rho$ into $d_\rho^{\hat{\pi}}$.
Now, let us consider the second case where constraints are violated.
In this case, the policy is updated as follows:
\begin{equation}
\label{eq: next policy parameter2}
\begin{aligned}
\theta_{t+1} &= \theta_t + \frac{\alpha_t}{1-\gamma}(\alpha_t \nu_t A_R^t - \sum_i \lambda_{t,i} A_{i, g_i}^t).
\end{aligned}
\end{equation}
As (\ref{eq: converegence proof 11}) derived in the first case,
\begin{equation}
\label{eq: converegence proof 21}
\begin{aligned}
&\alpha_t \nu_t (J_R(\theta_{t+1}) - J_R(\theta_t)) + \sum_i \lambda_{t,i}(J_{C_i}(\theta_t) -  J_{C_i}(\theta_{t+1})) \\
&= \frac{1}{1 - \gamma} \sum_{\bar{s}} d_\rho^{\pi_{t+1}}(\bar{s}) \sum_a \pi_{t+1}(a|\bar{s}) (\alpha_t \nu_t A_R^t(\bar{s}, a) - \sum_i \lambda_{t,i} A_{i, g_i}^t(\bar{s}, a)) \\
&= \frac{1}{1 - \gamma} \sum_{\bar{s}} d_\rho^{\pi_{t+1}}(\bar{s}) \sum_a (\pi_{t+1}(a|\bar{s}) - \pi_{t}(a|\bar{s})) (\alpha_t \nu_t A_R^t(\bar{s}, a) - \sum_i \lambda_{t,i} A_{i, g_i}^t(\bar{s}, a)) \\
&\leq \frac{2}{1 - \gamma} \sum_{\bar{s}} d_\rho^{\pi_{t+1}}(\bar{s}) \max_a \big|\alpha_t\nu_t A_R^t(\bar{s}, a) - \sum_i \lambda_{t,i} A_{i, g_i}^t(\bar{s}, a) \big| D_\mathrm{TV}(\pi_{t+1}(\cdot|\bar{s}), \pi_{t}(\cdot|\bar{s})) \\
&\leq \frac{2 ||\theta_{t+1} - \theta_{t}||_\infty}{1 - \gamma} \sum_{\bar{s}} d_\rho^{\pi_{t+1}}(\bar{s}) \max_a \big| \alpha_t \nu_t A_R^t(\bar{s}, a) - \sum_i \lambda_{t,i} A_{i, g_i}^t(\bar{s}, a) \big| \\
&\leq \frac{2\alpha_t}{(1-\gamma)^2} \big|\big| \alpha_t\nu_t A_R^t - \sum_i \lambda_{t,i} A_{i, g_i}^t \big|\big|_\infty^2 \leq 2 \alpha_t (\alpha_t \lambda_\mathrm{max} R_\mathrm{max} + N\lambda_\mathrm{max}C_\mathrm{max})^2/(1-\gamma)^4.
\end{aligned}
\end{equation}
Using (\ref{eq: next poliy exression}) and (\ref{eq: next policy parameter2}),
\begin{equation}
\label{eq: convergence proof 22}
\begin{aligned}
&\alpha_t \nu_t(J_R(\theta_{t+1}) - J_R(\theta_t)) + \sum_i \lambda_{t,i}(J_{C_i}(\theta_t) -  J_{C_i}(\theta_{t+1})) \\
&= \frac{1}{1-\gamma} \sum_{\bar{s}\in \bar{S}}d_\rho^{\pi_{t+1}}(\bar{s}) \sum_{a \in A}\pi_{t+1}(a|\bar{s}) (\alpha_t \nu_t A_R^t(\bar{s}, a) - \sum_i \lambda_{t,i} A_{i, g_i}^t(\bar{s}, a)) \\
&= \frac{1}{\alpha_t} \sum_{\bar{s}\in \bar{S}}d_\rho^{\pi_{t+1}}(\bar{s}) \sum_{a \in A}\pi_{t+1}(a|\bar{s}) \log (\pi_{t+1}(a|\bar{s}) Z_t(\bar{s})/\pi_t(a|\bar{s})) \\
&= \frac{1}{\alpha_t} \sum_{\bar{s}\in \bar{S}}d_\rho^{\pi_{t+1}}(\bar{s}) (D_\mathrm{KL}(\pi_{t+1}(\cdot|\bar{s}) || \pi_{t}(\cdot|\bar{s})) + \log Z_t(\bar{s})) \\
& \geq \frac{1}{\alpha_t} \sum_{\bar{s}\in \bar{S}}d_\rho^{\pi_{t+1}}(\bar{s}) \log Z_t(\bar{s}) \geq \frac{1 - \gamma}{\alpha_t} \sum_{\bar{s}\in \bar{S}}\rho(\bar{s}) \log Z_t(\bar{s}).
\end{aligned}
\end{equation}
By combining (\ref{eq: converegence proof 21}) and (\ref{eq: convergence proof 22}), the following is satisfied for any $\rho$:
\begin{equation}
\label{eq: convergence proof 23}
\begin{aligned}
\frac{1 - \gamma}{\alpha_t} \sum_{\bar{s}\in \bar{S}}\rho(\bar{s}) \log Z_t(\bar{s}) \leq \frac{2\alpha_t}{(1-\gamma)^4}(\alpha_t \lambda_\mathrm{max} R_\mathrm{max} + N\lambda_\mathrm{max}C_\mathrm{max})^2.
\end{aligned}
\end{equation}
As in (\ref{eq: unviolated case inequality}), the following is satisfied for any policy $\hat{\pi}$:
\begin{equation}
\label{eq: violated case inequality}
\begin{aligned}
&\alpha_t\nu_t (J_R(\hat{\pi}) - J_R(\pi_t)) + \sum_i \lambda_{t,i}(J_{C_i}(\pi_t) -  J_{C_i}(\hat{\pi})) \\
&= \frac{1}{1-\gamma} \sum_{\bar{s}\in \bar{S}}d_\rho^{\hat{\pi}}(\bar{s}) \sum_{a \in A}\hat{\pi}(a|\bar{s}) (\alpha_t\nu_t A_R^t(\bar{s}, a) - \sum_i \lambda_{t,i} A_{i, g_i}^t(\bar{s}, a)) \\
&= \frac{1}{\alpha_t} \sum_{\bar{s}\in \bar{S}}d_\rho^{\hat{\pi}}(\bar{s}) \sum_{a \in A}\hat{\pi}(a|\bar{s}) \log (\pi_{t+1}(a|\bar{s}) Z_t(\bar{s}) / \pi_t(a|\bar{s})) \\
&= \frac{1}{\alpha_t} \sum_{\bar{s}\in \bar{S}}d_\rho^{\hat{\pi}}(\bar{s}) (D_\mathrm{KL}(\hat{\pi}(\cdot|\bar{s}) || \pi_{t}(\cdot|\bar{s})) - D_\mathrm{KL}(\hat{\pi}(\cdot|\bar{s}) || \pi_{t+1}(\cdot|\bar{s})) + \log Z_t(\bar{s})) \\
& \leq \frac{1}{\alpha_t} \sum_{\bar{s}\in \bar{S}}d_\rho^{\hat{\pi}}(\bar{s}) (D_\mathrm{KL}(\hat{\pi}(\cdot|\bar{s}) || \pi_{t}(\cdot|\bar{s})) - D_\mathrm{KL}(\hat{\pi}(\cdot|\bar{s}) || \pi_{t+1}(\cdot|\bar{s}))) \\
&+ \frac{2\alpha_t}{(1-\gamma)^5}(\alpha_t \lambda_\mathrm{max} R_\mathrm{max} + N\lambda_\mathrm{max}C_\mathrm{max})^2.
\end{aligned}
\end{equation}
Now, the main inequalities for both cases have been derived.
Let us denote by $\mathcal{N}$ the set of time steps in which the constraints are not violated.
This means that if $t \in \mathcal{N}$, $J_{C_i}(\pi_t)\leq d_i \; \forall i$.
Then, using (\ref{eq: unviolated case inequality}) and (\ref{eq: violated case inequality}), we have:
\begin{equation}
\label{eq: convergence proof 40}
\begin{aligned}
&\sum_{t \in \mathcal{N}} \left( \alpha_t (J_R(\hat{\pi}) - J_R(\pi_t)) - \alpha_t^2\sum_i \lambda_{t,i}(J_{C_i}(\hat{\pi}) - J_{C_i}(\pi_t)) \right) \\
&\quad +\sum_{t \notin \mathcal{N}} \left(\alpha_t^2\nu_t(J_R(\hat{\pi}) - J_R(\pi)) - \alpha_t \sum_i \lambda_{t,i}(J_{C_i}(\hat{\pi}) - J_{C_i}(\pi_t)) \right) \\
&\leq \sum_{\bar{s}\in \bar{S}}d_\rho^{\hat{\pi}}(\bar{s}) D_\mathrm{KL}(\pi_f(\cdot|\bar{s}) || \pi_{0}(\cdot|\bar{s})) + \sum_{t\in \mathcal{N}} \frac{2\alpha_t^2}{(1-\gamma)^5}(R_\mathrm{max} + \alpha_t N \lambda_\mathrm{max} C_\mathrm{max})^2 \\
&\quad + \sum_{t \notin \mathcal{N}} \frac{2\alpha_t^2}{(1-\gamma)^5}(\alpha_t \lambda_\mathrm{max} R_\mathrm{max} + N \lambda_\mathrm{max} C_\mathrm{max})^2.
\end{aligned}
\end{equation}
If $t \notin \mathcal{N}$ and $\lambda_{t,i} > 0$, $J_{C_i}(\pi_t) > d_i$ and $\sum_i\lambda_{t,i} = 1$.
Thus, $\sum_i \lambda_{t,i}(J_{C_i}(\pi_f) - J_{C_i}(\pi_t)) \leq -\eta$.
By substituting $\hat{\pi}$ into $\pi_f$ in (\ref{eq: convergence proof 40}),
\begin{equation*}
\begin{aligned}
&\sum_{t \in \mathcal{N}} \left( \alpha_t (J_R(\pi_f) - J_R(\pi_t)) - \alpha_t^2\sum_i \lambda_{t,i}(J_{C_i}(\pi_f) - J_{C_i}(\pi_t)) \right) \\
&\quad + \sum_{t \notin \mathcal{N}}( \alpha_t\eta + \alpha_t^2\nu_t(J_R(\pi_f) - J_R(\pi)) ) \\
&\leq \sum_{\bar{s}\in \bar{S}}d_\rho^{\pi_f}(\bar{s}) D_\mathrm{KL}(\pi_f(\cdot|\bar{s}) || \pi_{0}(\cdot|\bar{s})) \\
&\quad + \sum_{t} \alpha_t^2\underbrace{\frac{2\max(\alpha_0, 1)^2}{(1-\gamma)^5}(\max(\lambda_\mathrm{max}, 1)R_\mathrm{max} + N \lambda_\mathrm{max} C_\mathrm{max})^2}_{=:K_1}.
\end{aligned}
\end{equation*}
For brevity, we denote $D_\mathrm{KL}(\pi_f||\pi_0) = \sum_{\bar{s}\in \bar{S}}d_\rho^{\pi_f}(\bar{s}) D_\mathrm{KL}(\pi_f(\cdot|\bar{s}) || \pi_{0}(\cdot|\bar{s}))$.
Since $\sum_{t\notin \mathcal{N}}\alpha_t = \sum_t \alpha_t - \sum_{t\in \mathcal{N}}\alpha_t$, (\ref{eq: convergence proof 40}) can be modified as follows:
\begin{equation*}
\begin{aligned}
&\sum_{t \in \mathcal{N}} \alpha_t (J_R(\pi_f) - J_R(\pi_t) - \eta) + \eta\sum_t \alpha_t \\
&\leq D_\mathrm{KL}(\pi_f || \pi_{0}) + K_1\sum_t \alpha_t^2 + \sum_{t\in\mathcal{N}}\left(\alpha_t^2\sum_i \lambda_{t,i}(J_{C_i}(\pi_f) - J_{C_i}(\pi_t)) \right) \\
&\quad - \sum_{t\notin\mathcal{N}}\left( \alpha_t^2\nu_t(J_R(\pi_f) - J_R(\pi)) \right) \\
&\leq D_\mathrm{KL}(\pi_f || \pi_{0}) + \sum_t \alpha_t^2 \underbrace{\left(K_1 + 2N\lambda_\mathrm{max}\frac{C_\mathrm{max}}{1-\gamma} + 2\lambda_\mathrm{max}\frac{R_\mathrm{max}}{1-\gamma} \right)}_{=:K_2}.
\end{aligned}
\end{equation*}
It can be simplified as follows:
\begin{equation*}
\sum_{t \in \mathcal{N}} \alpha_t (J_R(\pi_f) - J_R(\pi_t) - \eta) + \eta\sum_t \alpha_t \leq D_\mathrm{KL}(\pi_f || \pi_{0}) + K_2\sum_t \alpha_t^2.
\end{equation*}
If $\pi_0$ is set with positive values in all actions, $D_\mathrm{KL}(\hat{\pi}||\pi_0)$ is bounded for $\forall \hat{\pi}$.
Additionally, due to the Robbins-Monro condition, RHS converges to some real number as $T$ goes to infinity.
Since $\eta \sum_t \alpha_t$ in LHS goes to infinity, $\sum_{t\in \mathcal{N}} \alpha_t (J_R(\pi_f) - J_R(\pi_t) - \eta)$ must go to minus infinity.
It means that $\sum_{t\in\mathcal{N}}\alpha_t =\infty$ since $|J_R(\pi)| \leq R_\mathrm{max}/(1-\gamma)$.
By substituting $\hat{\pi}$ into $\pi^*$ in (\ref{eq: convergence proof 40}),
\begin{equation*}
\begin{aligned}
&\sum_{t \in \mathcal{N}} \alpha_t (J_R(\pi^*) - J_R(\pi_t))  - \sum_{t \notin \mathcal{N}} \alpha_t \left( \sum_i \lambda_{t,i}(J_{C_i}(\pi^*) - J_{C_i}(\pi_t)) \right) \\
&\leq D_\mathrm{KL}(\pi^* || \pi_{0}) + \sum_{t\in \mathcal{N}} \frac{2\alpha_t^2}{(1-\gamma)^5}(R_\mathrm{max} + \alpha_t N \lambda_\mathrm{max} C_\mathrm{max})^2 \\
&\quad + \sum_{t \notin \mathcal{N}} \frac{2\alpha_t^2}{(1-\gamma)^5}(\alpha_t \lambda_\mathrm{max} R_\mathrm{max} + N \lambda_\mathrm{max} C_\mathrm{max})^2 \\
&\quad + \sum_{t \in \mathcal{N}} \alpha_t^2\sum_i \lambda_{t,i}(J_{C_i}(\pi^*) - J_{C_i}(\pi_t)) - \sum_{t \notin \mathcal{N}} \alpha_t^2\nu_t(J_R(\pi^*) - J_R(\pi)) \\
&\leq D_\mathrm{KL}(\pi^* || \pi_{0}) + K_2 \sum_t \alpha_t^2.
\end{aligned}
\end{equation*}
Since $\sum_i \lambda_{t,i}(J_{C_i}(\pi^*) - J_{C_i}(\pi_t)) \leq 0$ for $t \notin \mathcal{N}$, the above equation is rewritten as follows:
\begin{equation*}
\sum_{t \in \mathcal{N}} \alpha_t (J_R(\pi^*) - J_R(\pi_t)) \leq D_\mathrm{KL}(\pi^* || \pi_{0}) + K_2 \sum_t \alpha_t^2.
\end{equation*}
Since $\sum_{t\in\mathcal{N}}\alpha_t =\infty$ and the Robbins-Monro condition is satisfied, the following must be held:
\begin{equation*}
J_R(\pi^*) - \lim_{t \to \infty} J_R(\pi_{\mathcal{N}[t]}) = 0,
\end{equation*}
where $\mathcal{N}[t]$ is the $t$th element of $\mathcal{N}$.
Consequently, the policy converges to an optimal policy.
\end{proof}

\section{Weight Selection Strategy}
\label{sec: lambda strategy}

In this section, we first identify $\nu_t$ and $\lambda_{t,i}$ for existing primal approach-based safe RL algorithms and present our strategy.
For the existing methods, we analyze several safe RL algorithms including constrained policy optimization (CPO, \citep{achiam2017constrained}), projection-based constrained policy optimization (PCPO, \citep{yang2020pcpo}), constraint-rectified policy optimization (CRPO, \citep{xu2021crpo}), penalized proximal policy optimization (P3O, \citep{zhang2022penalized}), interior-point policy optimization (IPO, \citep{liu2020ipo}), and safe distributional actor-critic (SDAC, \citep{kim2023sdac}).
Since the policy gradient is obtained differently depending on whether the constraints are satisfied (CS) or violated (CV), we analyze the existing methods for each of the two cases.

\textbf{Case 1: Constraints are satisfied (CS).}
First, CPO \citep{achiam2017constrained} and SDAC \citep{kim2023sdac} find a direction of the policy gradient by solving the following linear programming with linear and quadratic constraints (LQCLP):
\begin{equation*}
\max_g \nabla J_R(\theta_t)^Tg \; \mathbf{s.t.} \nabla J_{C_i}(\theta_t)^T g + J_{C_i}(\theta_t) \leq d_i \; \forall i, \; \frac{1}{2}g^T F(\theta_t)g \leq \epsilon,
\end{equation*}
where $\epsilon$ is a trust region size.
The LQCLP can be addressed by solving the following dual problem:
\begin{equation*}
\begin{aligned}
\lambda^*, \nu^* = \underset{\lambda \geq 0, \nu \geq 0}{\arg \max} &- \nabla J_R(\theta_t)^Tg^*(\lambda, \nu) + \sum_{i=1}^N \lambda_i (\nabla J_{C_i}(\theta_t)^T g^*(\lambda, \nu) + J_{C_i}(\theta_t) - d_i) \\
&+ \nu(\frac{1}{2}g^*(\lambda, \nu)^T F(\theta_t) g^*(\lambda, \nu) - \epsilon),
\end{aligned}
\end{equation*}
where $g^*(\lambda, \nu) = \frac{1}{\nu}F^\dagger(\theta_t) \left( \nabla J_R(\theta_t) - \sum_i \lambda_i \nabla J_{C_i}(\theta_t) \right)$.
Then, the policy is updated in the following direction:
\begin{equation}
\label{eq: weight stategy SDAC}
g_t = F^\dagger(\theta_t)(\nabla J_R(\theta_t) - \alpha_t \sum_i \min(\lambda_i^*/\alpha_t, \lambda_\mathrm{max}) \nabla J_{C_i}(\theta_t)),
\end{equation}
which results in $\lambda_{t,i} = \min(\lambda_i^*/\alpha_t, \lambda_\mathrm{max})$.
P3O \citep{zhang2022penalized}, PCPO \citep{yang2020pcpo}, and CRPO \citep{xu2021crpo} update the policy only using the objective function in this case as follows:
\begin{equation*}
g_t = F^\dagger(\theta_t)\nabla J_R(\theta_t) \Rightarrow \lambda_{t,i} = 0.
\end{equation*}
IPO \citep{liu2020ipo} update a policy in the following direction:
\begin{equation*}
g_t = F^\dagger(\theta_t) \left( \nabla J_R(\theta_t) - \kappa \sum_i \nabla J_{C_i}(\theta_t)/(d_i - J_{C_i}(\theta_t)) \right),
\end{equation*}
where $\kappa$ is a penalty coefficient.
As a result, $\lambda_{t,i}$ is computed as $\min(\kappa/((d_i -J_{C_i}(\theta_t))\alpha_t), \lambda_\mathrm{max})$.

\textbf{Case 2: Constraints are violated (CV).}
CPO, PCPO and IPO did not handle the constraint violation in multiple constraint settings, so we exclude them in this case.
First, SDAC finds a recovery direction by solving the following quadratic programming (QP):
\begin{equation*}
\min_g \frac{1}{2}g^T F(\theta_t)g \; \mathbf{s.t.} \nabla J_{C_i}(\theta_t)^T g + J_{C_i}(\theta_t) \leq d_i \; \text{for} \; i\in \{i|J_{C_i}(\theta_t) > d_i\},
\end{equation*}
where we will denote $\{i|J_{C_i}(\theta_t) > d_i\}$ by $I_\mathrm{CV}$.
The QP can be addressed by solving the following dual problem:
\begin{equation*}
\lambda^* = \underset{\lambda \geq 0}{\arg\max} \frac{1}{2} g^*(\lambda)^T F(\theta_t) g^*(\lambda) + \sum_{i \in I_\mathrm{CV}} \lambda_i (\nabla J_{C_i}(\theta_t)^T g^*(\lambda) + J_{C_i}(\theta_t) - d_i),
\end{equation*}
where $g^*(\lambda) = - F^\dagger(\theta_t) \sum_{i\in I_\mathrm{CV}} \lambda_i \nabla J_{C_i}(\theta_t)$.
Then, the policy is updated in the following direction:
\begin{equation*}
g_t = F^\dagger(\theta_t) \left(\sum_{i \in I_\mathrm{CV}} -\frac{\lambda_i^*}{\sum_{j \in I_\mathrm{CV}}\lambda_j^*} \nabla J_{C_i}(\theta_t) \right) \Rightarrow \nu_t=0, \; \lambda_{t,i} = \frac{\lambda_i^*}{\sum_{j\in I_\mathrm{CV}}\lambda_j^*} \; \text{if} \; i\in I_\mathrm{CV} \; \text{else} \; 0.
\end{equation*}
If there is only a single constraint, PCPO is compatible with SDAC.
Next, CRPO first selects a violated constraint, whose index denoted as $k$, and calculates the policy gradient to minimize the selected constraint as follows:
\begin{equation*}
g_t = F^\dagger(\theta_t) \left(-\nabla J_{C_k}(\theta_t) \right) \Rightarrow \nu_t=0, \; \lambda_{t,i} = 1 \; \text{if} \; i=k \; \text{else} \; 0.
\end{equation*}
If there is only a single constraint, CPO is compatible with CRPO.
Finally, P3O update the policy in the following direction:
\begin{equation*}
g_t = F^\dagger(\theta_t)\left(\nabla J_R(\theta_t) - \kappa \sum_{i \in I_\mathrm{CV}} \nabla J_{C_i}(\theta_t) \right),
\end{equation*}
which results in $\nu_t = \min(1/(\alpha_t \kappa |I_\mathrm{CV}|), \lambda_\mathrm{max})$, and $\lambda_{t,i} = 1/|I_\mathrm{CV}|$ if $i \in I_\mathrm{CV}$ else $0$.

\textbf{Implementation of trust region method.}
To this point, we have obtained the direction of the policy gradient $g_t$, allowing the policy to be updated to $\theta_t + \alpha_t g_t$.
If we want to update the policy using a trust region method, as introduced in TRPO \citep{schulman2015trust}, we can adjust the learning rate $\alpha_t$ as follows:
\begin{equation}
\label{eq: trust region alpha}
\alpha_t = \frac{\epsilon_t}{\mathrm{Clip}(\sqrt{g_t^T F(\theta_t)g_t}, g_\mathrm{min}, g_\mathrm{max})},
\end{equation}
where $\epsilon_t$ is a trust region size following the Robbins-Monro condition, $g_\mathrm{min}$ and $g_\mathrm{max}$ are hyperparamters, and $\mathrm{Clip}(x, a, b) = \min(\max(x, a), b)$.
By adjusting the learning rate, the policy can be updated mostly within the trust region, defined as $\{\theta_t + g|g^T F(\theta_t)g \leq \epsilon_t^2\}$, while the learning rate still satisfies the Robbins-Monro condition.
Accordingly, we also need to modify $\lambda_{t,i}$ for the CS case and $\nu_t$ for the CV case, as they are  expressed with the learning rate $\alpha_t$.
We explain this modification process by using SDAC as an example.
The direction of the policy gradient of SDAC for the CS case expressed as follows:
\begin{equation*}
g_t = F^\dagger(\theta_t)(\nabla J_R(\theta_t) - \epsilon_t \sum_i \min(\lambda_i^*/\epsilon_t, \lambda_\mathrm{max}) \nabla J_{C_i}(\theta_t)),
\end{equation*}
where $\alpha_t$ in (\ref{eq: weight stategy SDAC}) is replaced with $\epsilon_t$.
The learning rate $\alpha_t$ is then calculated using $g_t$ and (\ref{eq: trust region alpha}).
Now, we need to express $g_t$ as $F^\dagger(\theta_t)(\nabla J_R(\theta_t) - \alpha_t \sum_i \lambda_{t,i} \nabla J_{C_i}(\theta_t))$. 
To this end, we can use $\alpha_t = \epsilon_t/C$, where $C=\mathrm{Clip}(\sqrt{g_t^T F(\theta_t)g_t}, g_\mathrm{min}, g_\mathrm{max})$.
Consequently, $\lambda_{t,i}$ is expressed as follows:
\begin{equation*}
\lambda_{t,i} = C \min(\lambda_i^*/\epsilon_t, \lambda_\mathrm{max}),
\end{equation*}
and the maximum of $\lambda_{t,i}$ and $\nu_t$ is adjusted to $g_\mathrm{max} \lambda_\mathrm{max}$.

\textbf{Proposed strategy.}
Now, we introduce the proposed strategy for determining $\lambda_{t,i}$ and $\nu_t$.
We first consider the CV case.
\citet{kim2023sdac} empirically showed that processing multiple constraints simultaneously at each update step effectively projects the current policy onto a feasible set of policies. 
Therefore, we decide to use the same strategy as SDAC, which deals with constraints simultaneously, and it is written as follows:
\begin{equation*}
\nu_t=0, \; \lambda_{t,i} = \begin{cases}
\lambda_i^*/\left(\sum_{j\in I_\mathrm{CV}}\lambda_j^*\right)  & \text{if} \; i\in I_\mathrm{CV}, \\
0 & \text{otherwise,}
\end{cases}
\end{equation*}
where $\lambda^* = {\arg\max}_{\lambda \geq 0} \frac{1}{2} g^*(\lambda)^T F(\theta_t) g^*(\lambda) + \sum_{i \in I_\mathrm{CV}} \lambda_i \left(\nabla J_{C_i}(\theta_t)^T g^*(\lambda) + J_{C_i}(\theta_t) - d_i\right)$ and $g^*(\lambda) = - F^\dagger(\theta_t) \sum_{i\in I_\mathrm{CV}} \lambda_i \nabla J_{C_i}(\theta_t)$.
Next, let us consider the CS case.
We hypothesize that concurrently addressing both constraints and objectives when calculating the policy gradient can lead to more stable constraint satisfaction. 
Therefore, CPO, SDAC, and IPO can be suitable candidates.
However, CPO and SDAC methods require solving the LQCLP whose dual problem is nonlinear, and IPO can introduce computational errors when $J_{C_i}(\theta_t)$ is close to $d_i$.
To address these challenges, we propose to calculate the policy gradient direction by solving a QP instead of an LQCLP.
By applying a concept of transforming objectives into constraints \citep{kim2024scale} to the safe RL problem, we can formulate the following QP:
\begin{equation*}
\min_g \frac{1}{2}g^T F(\theta_t)g \; \mathbf{s.t.} \; \nabla J_R(\theta_t)^T g \geq e, \; \nabla J_{C_i}(\theta_t)^T g + J_{C_i}(\theta_t) \leq d_i \; \forall i,
\end{equation*}
where $e:=\epsilon_t \sqrt{\nabla J_R(\theta_t)^T F^\dagger(\theta_t) \nabla J_R(\theta_t)}$.
According to \citep{kim2024scale}, $e$ represents the maximum value of $\nabla J_R(\theta_t)^T g$ when $g$ is within the trust region, $\{g|g^T F(\theta_t)g \leq \epsilon_t^2\}$.
Therefore, the transformed constraint, $\nabla J_R(\theta_t)^T g \geq e$, has the same effect of improving the objective function $J_R(\theta)$ within the trust region.
The QP is then addressed by solving the following dual problem:
\begin{equation*}
\begin{aligned}
\lambda^*, \nu^* = \underset{\lambda \geq 0, \nu \geq 0}{\arg \max} & \frac{1}{2} g^*(\lambda, \nu)^T F(\theta_t)g^*(\lambda, \nu) + \nu(e - \nabla J_R(\theta_t)^T g^*(\lambda, \nu)) \\
&+ \sum_{i=1}^N \lambda_i (\nabla J_{C_i}(\theta_t)^T g^*(\lambda, \nu) + J_{C_i}(\theta_t) - d_i), \\
\end{aligned}
\end{equation*}
where $g^*(\lambda, \nu) = F^\dagger(\theta_t)(\nu \nabla J_{R}(\theta_t) - \sum_{i\in I_\mathrm{CV}} \lambda_i \nabla J_{C_i}(\theta_t))$.
Then, the policy is updated in the following direction:
\begin{equation*}
g_t = F^\dagger(\theta_t)\left(\nabla J_R(\theta_t) - \epsilon_t \sum_i \min\left(\frac{\lambda_i^*}{\nu^* \epsilon_t}, \lambda_\mathrm{max}\right) \nabla J_{C_i}(\theta_t)\right),
\end{equation*}
and the learning rate is calculated using (\ref{eq: trust region alpha}) to ensure that the policy is updated within the trust region.
Finally, $\lambda_{t,i}$ is expressed as $C\min(\lambda_i^*/(\nu^* \epsilon_t), \lambda_\mathrm{max})$, where $C=\mathrm{Clip}(\sqrt{g_t^T F(\theta_t)g_t}, g_\mathrm{min}, g_\mathrm{max})$.

\section{Implementation Details}
\label{sec: implementation detail}

\begin{algorithm}[tbh]
\caption{Spectral-Risk-Constrained Policy Optimization (SRCPO)}
\label{algo: practical implementation}
\begin{algorithmic}
\STATE {\bfseries Input:} Policy parameter $\theta$, sampler parameter $\phi$, critic parameter $\psi$, and replay buffer $D$.
\FOR{epochs $=1$ {\bfseries to} $P$}
    \FOR{episodes $=1$ {\bfseries to} $E$}
        \IF{$u \leq \epsilon$, where $u \sim U[0, 1]$,}
            \STATE Sample $\boldsymbol{\beta}$ from a uniform distribution: $\beta_i[j] \sim U(\beta_i[j-1], C_\mathrm{max}/(1-\gamma))$ $\forall i$ and $\forall j$, where $\beta_i[0]=0$.
        \ELSE
            \STATE Sample $\boldsymbol{\beta} \sim \xi_\phi$.
        \ENDIF
        \STATE Collect a trajectory $\tau=\{\bar{s}_t, a_t, r_t, \{c_{i, t}\}_{i=1}^N\}_{t=0}^T$ using $\pi_\theta(a|\bar{s}; \boldsymbol{\beta})$, and store $(\boldsymbol{\beta}, \tau)$ in $D$.
    \ENDFOR
    \STATE Set $G^\theta=\{\}$ and $G^\phi = \{\}$.
    \FOR{updates $=1$ {\bfseries to} $U$}
        \STATE Sample $(\boldsymbol{\beta}, \tau) \sim D$.
        \STATE Update the critics, $Z_{R,\psi}(\bar{s}, a; \boldsymbol{\beta})$ and $Z_{C_i,\psi}(\bar{s}, a; \boldsymbol{\beta})$, using the quantile regression loss.
        \STATE Calculate the policy gradient of $\pi_\theta(a|\bar{s}; \boldsymbol{\beta})$ using the update rule, described in Section \ref{sec: policy update rule}, and store it to $G^\theta$.
        \STATE Calculate the gradient of the sampler $\xi_\phi$ using the update rule, described in (\ref{eq: sampler update}), and store it to $G^\phi$.
    \ENDFOR
    \STATE Update the policy parameter by the average of $G^\theta$.
    \STATE Update the sampler parameter by the average of $G^\phi$.
\ENDFOR
\STATE {\bfseries Output:} $\pi_\theta(\cdot|\cdot; \boldsymbol{\beta})$, where $\boldsymbol{\beta} \sim \xi_\phi$.
\end{algorithmic}
\end{algorithm}

In this section, we present a practical algorithm that utilizes function approximators, such as neural networks, to tackle more complex tasks. 
The comprehensive procedure is presented in Algorithm \ref{algo: practical implementation}.

First, it is necessary to train individual policies for various $\boldsymbol{\beta}$, so we model a policy network to be conditioned on $\boldsymbol{\beta}$, expressed as $\pi_\theta(a|\bar{s}; \boldsymbol{\beta})$.
This network structure enables the implementation of policies with various behaviors by conditioning the policy on different $\boldsymbol{\beta}$ values. 
In order to train this $\boldsymbol{\beta}$-conditioned policy across a broad range of the $\boldsymbol{\beta}$ space, a $\epsilon$-greedy strategy can be used. 
At the beginning of each episode, $\boldsymbol{\beta}$ is sampled uniformly with a probability of $\epsilon$; otherwise, it is sampled using the sampler $\xi_\phi$.
In our implementation, we did not use the uniform sampling process by setting $\epsilon=0$ to reduce fine-tuning efforts, but it can still be used to enhance practical performance. 

After collecting rollouts with the sampled $\boldsymbol{\beta}$, the critic, policy, and sampler need to be updated. 
For the critic, a distributional target is calculated using the TD($\lambda$) method \citep{kim2023sdac}, and we update the critic networks to minimize the quantile regression loss \citep{dabney2018quantile}, defined as follows:
\begin{equation*}
\mathcal{L}(\psi) := \sum_{l=1}^L\underset{(\bar{s},a,\boldsymbol{\beta}) \sim D}{\mathbb{E}} \left[ \underset{z \sim \hat{Z}_{C_i}(\bar{s},a; \boldsymbol{\beta})}{\mathbb{E}} \bigg[ \rho_{\frac{2l-1}{2L}} \Big( z - Z_{C_i, \psi}(\bar{s},a; \boldsymbol{\beta})[l] \Big) \bigg] \right],
\end{equation*}
where $\rho_l(u):=u\cdot(l-\mathbf{1}_{u<0})$, and $\hat{Z}_{C_i}(\bar{s},a; \boldsymbol{\beta})$ is the target distribution of $Z_{C_i, \psi}(\bar{s},a; \boldsymbol{\beta})$.
The policy gradient for a specific $\boldsymbol{\beta}$ can be calculated according to the update rule described in Section \ref{sec: policy update rule}. 
Consequently, the policy is updated using the policy gradient calculated with $\boldsymbol{\beta}$ and rollouts, which are extracted from the replay buffer. 
Finally, the gradient of the sampler for a given $\boldsymbol{\beta}$ can be calculated as in (\ref{eq: sampler update}), so the sampler is updated with the average of gradients, which are calculated across multiple $\boldsymbol{\beta}$.

\section{Experimental Details}
\label{sec: experimental details}

In this section, we provide details on tasks, network structure, and hyperparameter settings.

\begin{figure}[!t]
    \centering
    \begin{subfigure}[b]{0.24\textwidth}
        \centering
        \includegraphics[width=\textwidth]{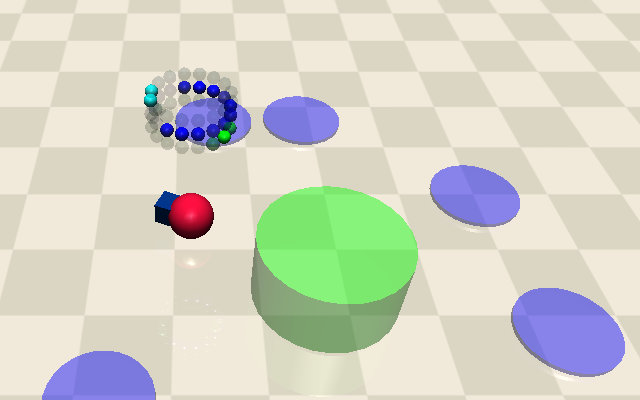}
        \caption{Point goal task}
    \end{subfigure}
    \hfill
    \begin{subfigure}[b]{0.24\textwidth}
        \centering
        \includegraphics[width=\textwidth]{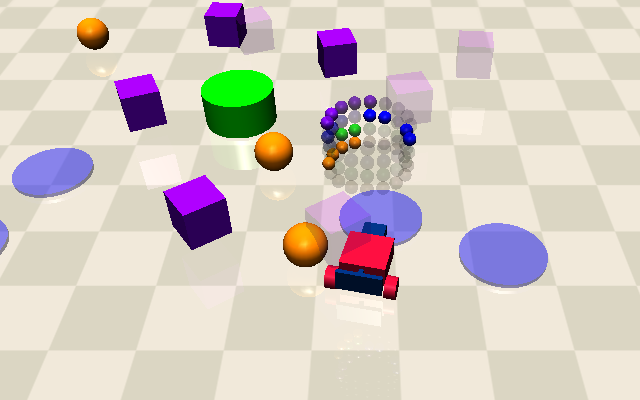}
        \caption{Car button task}
    \end{subfigure}
    \hfill
    \begin{subfigure}[b]{0.24\textwidth}
        \centering
        \includegraphics[width=\textwidth]{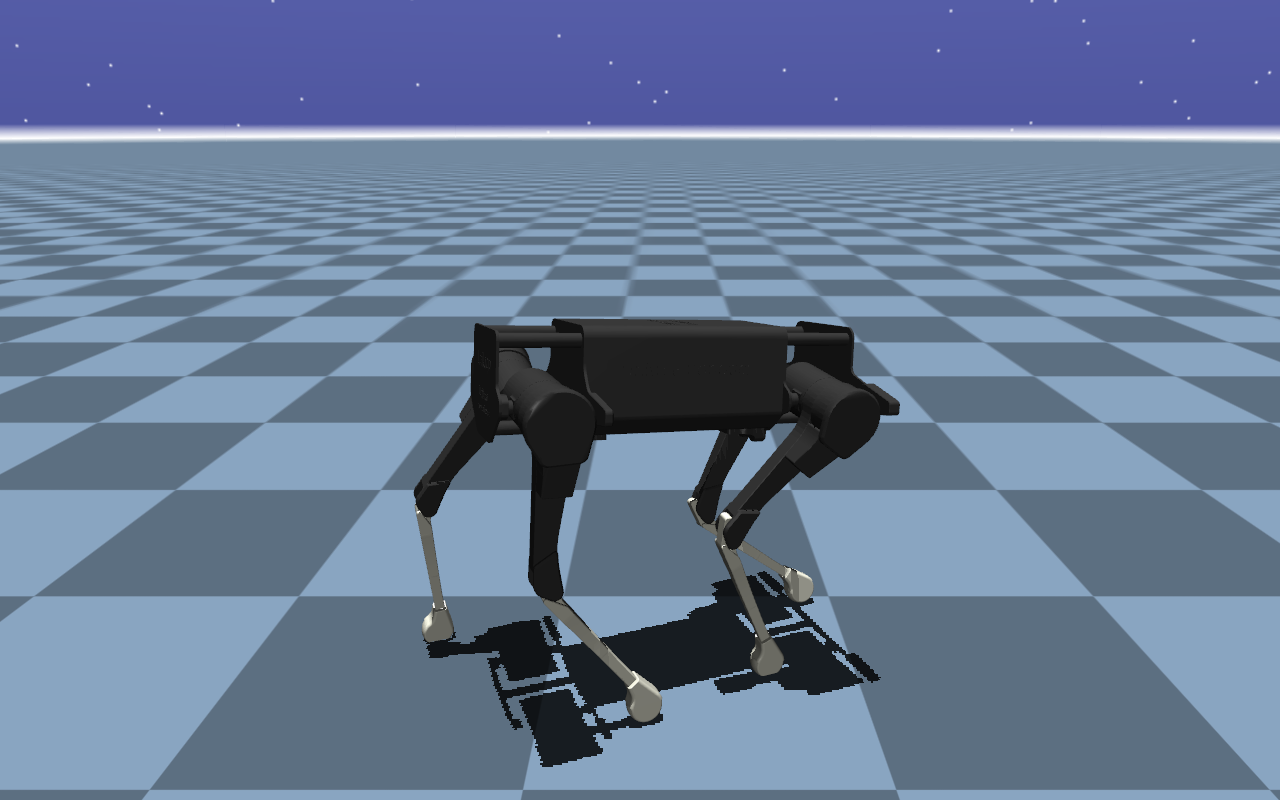}
        \caption{Quadrupedal robot}
    \end{subfigure}
    \hfill
    \begin{subfigure}[b]{0.24\textwidth}
        \centering
        \includegraphics[width=\textwidth]{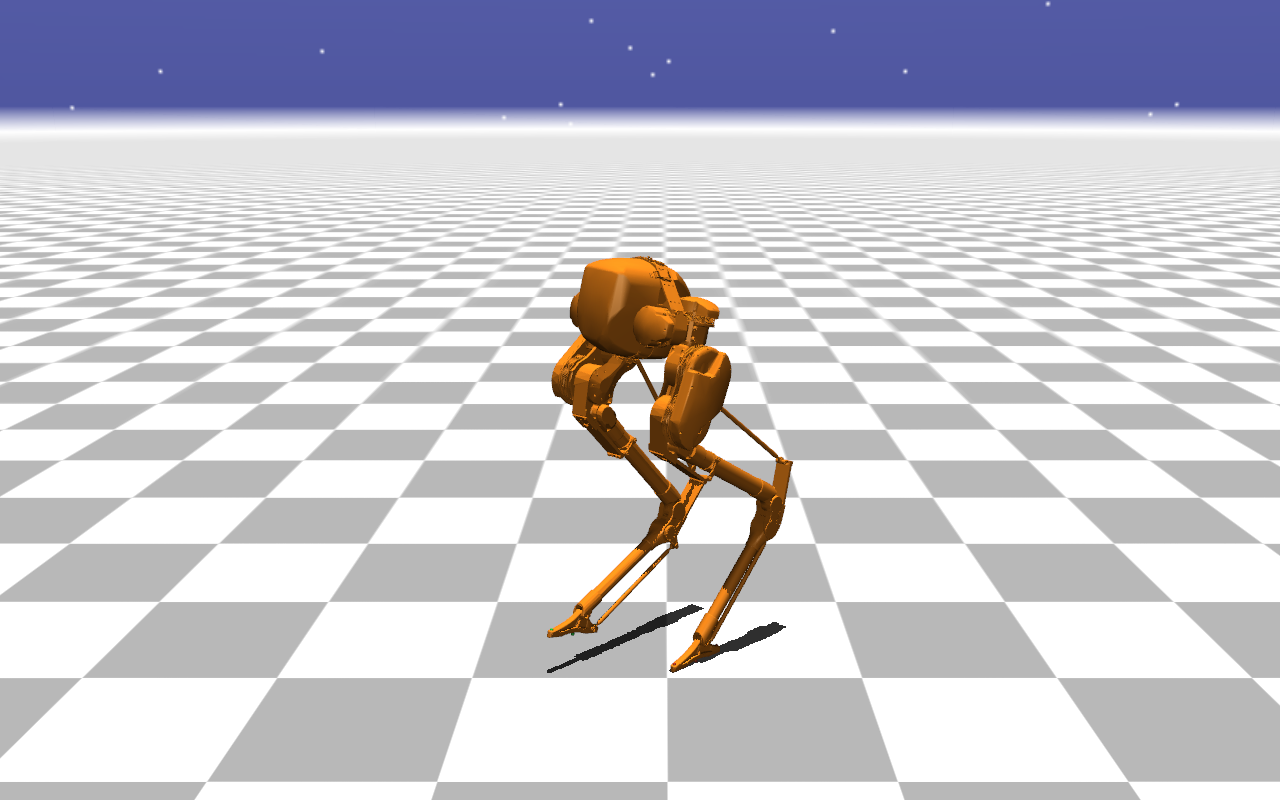}
        \caption{Bipedal robot}
    \end{subfigure}
    \caption{
    Rendered images of the Safety Gymnasium and the legged robot locomotion tasks.
    }
    \label{fig: snapshot of tasks}
\end{figure}

\subsection{Task Description}

There are two environments, legged robot locomotion \citep{kim2023sdac} and Safety Gymnasium \citep{ji2023safetygym}, and whose rendered images are presented in Figure \ref{fig: snapshot of tasks}.

\textbf{Legged Robot Locomotion.}
The legged robot locomotion tasks aim to control bipedal and quadrupedal robots to maintain a specified target velocity, which consists of linear velocity in the $X$ and $Y$ directions and rotational velocity in the $Z$ direction, without falling. 
Each robot takes actions as PD target joint positions at a frequency of 50 Hz, and a PD controller in each joint operates at 500 Hz. 
The dimension of the action space corresponds to the number of motors: ten for the bipedal robot and 12 for the quadrupedal robot.
The state space includes the command, gravity vector, current joint positions and velocities, foot contact phase, and a history of the joint positions and velocities. 
The reward function is defined as the negative of the squared Euclidean distance between the current velocity and the target velocity.
There are three cost functions to prevent robots from falling.
The first cost is to balance the robot, which is formulated as $\mathbf{1}_{g[2]/||g||_2 \geq \cos(\pi/12)}$, where $g$ is the gravity vector.
This function is activated when the base frame of the robot is tilted more than $15$ degrees.
The second cost is to maintain the height of the robot above a predefined threshold, which is expressed as $\mathbf{1}_{h \leq b}$, where $h$ is the height of the robot, and $b$ is set as $0.7$ in the bipedal robot and $0.35$ in the quadrupedal robot.
The third cost is to match the foot contact state with the desired timing. 
It is defined as the average number of feet that do not touch the floor at the desired time.
We set the thresholds for these cost functions as $0.025/(1-\gamma), 0.025/(1-\gamma)$, and $0.4/(1-\gamma)$, respectively, where $(1-\gamma)$ converts the threshold value from an average time horizon scale to an infinite time horizon scale.

\textbf{Safety Gymnasium.}
We use the goal and button tasks in the Safety Gymnasium \citep{ji2023safetygym}, and both tasks employ point and car robots.
The goal task is to control the robot to navigate to a target goal position without passing hazard area.
The button task is to control the robot to push a designated button among several buttons without passing hazard area or hitting undesignated buttons.
The dimension of the action space is two for all tasks. The state space dimensions are 60, 72, 76, and 88 for the point goal, car goal, point button, and car button tasks, respectively.
For more details on the action and state, please refer to \citep{ji2023safetygym}.
The reward function is defined to reduce the distance between the robot and either the goal or the designated button, with a bonus awarded at the moment of task completion. 
When the robot traverses a hazard area or collides with an obstacle, it incurs a cost of one, and the threshold for this cost is set as $0.025/(1-\gamma)$.

\subsection{Network Structure}

Across all algorithms, we use the same network architecture for the policy and critics. 
The policy network is structured to output the mean and standard deviation of a normal distribution, which is then squashed using the $\mathrm{Tanh}$ function as done in SAC \citep{haarnoja2018sac}. 
The critic network is based on the quantile distributional critic \citep{dabney2018quantile}. 
Details of these structures are presented in Table \ref{tab: network structure}.
For the sampler $\xi_\phi$, we define a trainable variable $\phi \in \mathbb{R}^{N\times (M-1)}$ and represent the mean of the truncated normal distribution as $\mu_{i, \phi}[j] = \exp(\phi[i, j])$.
To reduce fine-tuning efforts, we fix the standard deviation of the truncated normal distribution to $0.05$.
\begin{table}[!htb]
\caption{Details of network structures.}
\label{tab: network structure}
\centering
\resizebox{0.5\textwidth}{!}{%
\begin{tabular}{c|c|c}
\toprule
 & Parameter& Value \\
\midrule
 Policy network&Hidden layer& (512, 512)\\
 & Activation & LeakyReLU \\
 & Last activation & Linear \\
\midrule
 Reward critic network& Hidden layer&(512, 512)\\
 & Activation & LeakyReLU \\
 & Last activation & Linear \\
 & \texttt{\#} of quantiles & 25\\
 & \texttt{\#} of ensembles & 2\\
 \midrule
 Cost critic network& Hidden layer & (512, 512)\\
 & Activation & LeakyReLU \\
 & Last activation & SoftPlus \\
 & \texttt{\#} of quantiles & 25\\
 & \texttt{\#} of ensembles & 2\\
 \bottomrule
\end{tabular}%
}
\end{table}

\subsection{Hyperparameter Settings}

The hyperparameter settings for all algorithms applied to all tasks are summarized in Table \ref{tab: hyperparameter}.

\begin{table}[!htb]
\caption{Description on hyperparameter settings.}
\label{tab: hyperparameter}
\centering
\resizebox{1.0\textwidth}{!}{%
\begin{tabular}{c|c|c|c|c|c|c}
\toprule
 & SRCPO (Ours) & CPPO \citep{ying2022cppo} & CVaR-CPO \citep{zhang2024cvar} & SDAC \citep{kim2023sdac} & SDPO \citep{zhang2022sdpo} & WCSAC-D \citep{yang2023wcsac}\\
 \midrule
 Discount factor& $0.99$& $0.99$& $0.99$& $0.99$& $0.99$&$0.99$\\
 Policy learning rate&-&$3 \cdot 10^{-5}$& $3 \cdot 10^{-5}$& -& $3 \cdot 10^{-5}$&$3 \cdot 10^{-4}$\\
 Critic learning rate& $3 \cdot 10^{-4}$& $3 \cdot 10^{-4}$& $3 \cdot 10^{-4}$& $3 \cdot 10^{-4}$& $3 \cdot 10^{-4}$&$3 \cdot 10^{-4}$\\
 \texttt{\#} of target quantiles& $50$& $50$& $50$& $50$& $50$&-\\
 Coefficient of TD($\lambda$)& $0.97$& $0.97$& $0.97$& $0.97$& $0.97$&-\\
 Soft update ratio& -& -& -& -& -&$0.995$\\
 Batch size& $10000$& $5000$& $5000$& $10000$& $5000$&$256$\\
 Steps per update& $1000$& $5000$& $5000$& $1000$& $5000$&$100$\\
 \makecell{\texttt{\#} of policy \\update iterations}& $10$& $20$& $20$& -& $20$&$10$\\
 \makecell{\texttt{\#} of critic \\update iterations}& $40$& $40$& $40$& $40$& $40$&$10$\\
 Trust region size& $0.001$& $0.001$& $0.001$& $0.001$& $0.001$&-\\
 Line search decay& -& -& -& $0.8$& -&-\\
 \makecell{PPO Clip ratio \citep{schulman2017ppo}}& -& $0.2$& $0.2$& -& $0.2$&-\\
 \makecell{Length of \\replay buffer}& $100000$& $5000$& $5000$& $100000$& $5000$&$1000000$\\
 \makecell{Learning rate for \\auxiliary variables}& -& $0.05$& $0.01$& -& -&-\\
 \makecell{Learning rate for \\Lagrange multipliers}& -& $0.05$& $0.01$& -& -&$0.001$\\
 Log penalty coefficient& -& -& -& -& $5.0$&-\\
 Entropy threshold& -& -& -& -& -&$-|A|$\\
 Entropy learning rate& -& -& -& -& -&$0.001$\\
 Entropy coefficient& & $0.001$& -& -& $0.001$&-\\
 Sampler learning rate& $1\cdot 10^{-3}$&-& -& -& -&-\\
 $K$ for sampler target & $10$&-& -& -& -&-\\
 \bottomrule
\end{tabular}%
}
\end{table}

\newpage
\section{Additional Experimental Results}
\label{sec: additional experimental results}

\subsection{Computational Resources}

All experiments were conducted on a PC whose CPU and GPU are an Intel Xeon E5-2680 and NVIDIA TITAN Xp, respectively.
The training time for each algorithm on the point goal task is reported in Table \ref{tab: computation time}.
\begin{table}[h]
    \caption{Training time for the point goal task averaged over five runs.}
    \label{tab: computation time}
    \centering
    \resizebox{1.0\textwidth}{!}{%
    \begin{tabular}{c|c|c|c|c|c|c}
        \toprule
           &  SRCPO (Ours)&  CPPO& CVaR-CPO& SDAC&WCSAC-D&SDPO\\
        \midrule
        \makecell{Averaged\\training time}& 9h 51m 54s& 6h 17m 16s& 4h 43m 38s& 8h 46m 8s&14h 18m 20s&6h 50m 39s\\
        \bottomrule
    \end{tabular}%
    }
\end{table}

\subsection{Experiments on Various Risk Measures}

We have conducted studies on various risk measures in the main text.
In this section, we first describe the definition of the Wang risk measure \citep{wang2000class}, show the visualization of the discretization process, and finally present the training curves of the experiments conducted in Section \ref{sec: experiments on other risk measure}.

The Wang risk measure is defined as follows \citep{wang2000class}:
\begin{equation*}
\mathrm{Wang}_\alpha(X) := \int_0^1 F_X^{-1}(u) d\zeta_\alpha(u), \quad \text{where} \; \zeta_\alpha(u) := \Phi(\Phi^{-1}(u) - \alpha),
\end{equation*}
$\Phi$ is the cumulative distribution function (CDF) of the standard normal distribution, and $\zeta$ is called a distortion function.
It can be expressed in the form of the spectral risk measure as follows:
\begin{equation*}
\mathrm{Wang}_\alpha(X) = \int_0^1 F_X^{-1}(u) \sigma_\alpha(u) du, \quad \text{where} \; \sigma_\alpha(u) := \frac{\phi(\Phi^{-1}(u) - \alpha)}{\phi(\Phi^{-1}(u))},
\end{equation*}
and $\phi$ is the pdf of the standard normal distribution.
Since $\sigma_\alpha(u)$ goes to infinity when $u$ is approaching one, $\sigma_\alpha(u)$ at $u=1$ is not defined.
Thus, it is not a spectral risk measure.
Nevertheless, our parameterization method introduced in Section \ref{sec: discretization} can project $\mathrm{Wang}_\alpha$ onto a discretized class of spectral risk measures.

\begin{figure}[!thb]
    \centering
    \begin{subfigure}[b]{0.301\textwidth}
        \centering
        \includegraphics[width=\textwidth]{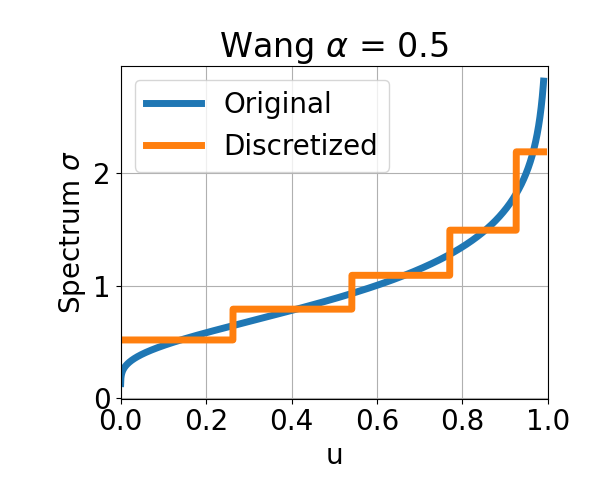}
    \end{subfigure}
    \hfill
    \begin{subfigure}[b]{0.301\textwidth}
        \centering
        \includegraphics[width=\textwidth]{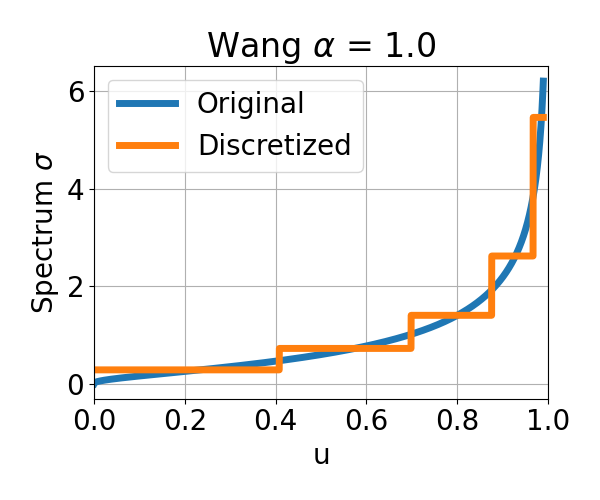}
    \end{subfigure}
    \hfill
    \begin{subfigure}[b]{0.301\textwidth}
        \centering
        \includegraphics[width=\textwidth]{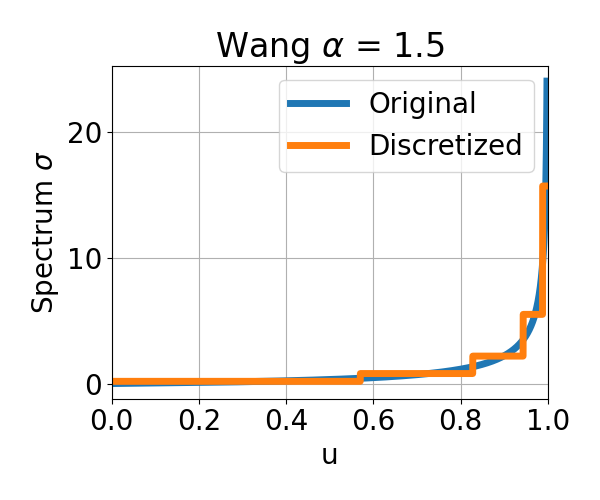}
    \end{subfigure}
    \caption{
    Visualization of the spectrum function of the Wang risk measure and the discretized results.
    }
    \label{fig: wang visualization}
\end{figure}

\begin{figure}[!thb]
    \centering
    \begin{subfigure}[b]{0.301\textwidth}
        \centering
        \includegraphics[width=\textwidth]{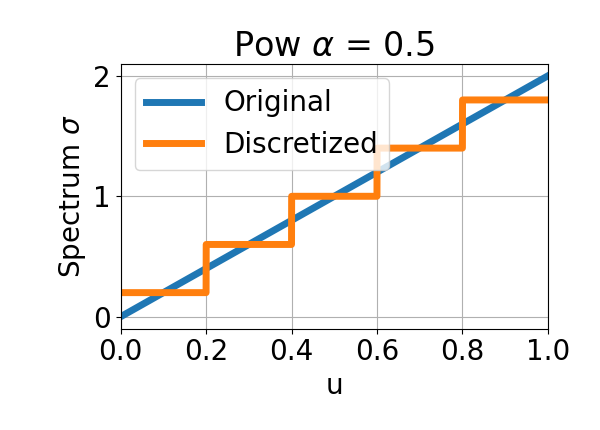}
    \end{subfigure}
    \hfill
    \begin{subfigure}[b]{0.301\textwidth}
        \centering
        \includegraphics[width=\textwidth]{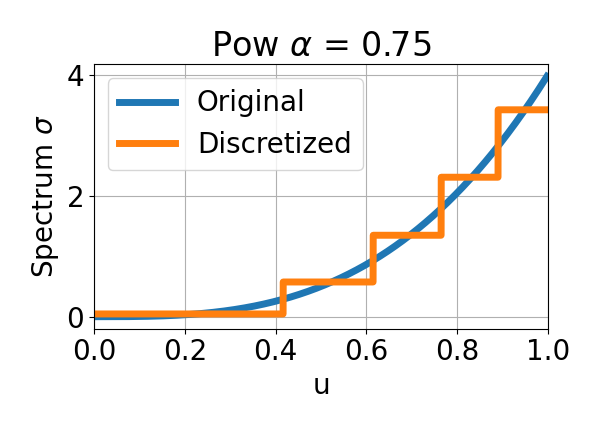}
    \end{subfigure}
    \hfill
    \begin{subfigure}[b]{0.301\textwidth}
        \centering
        \includegraphics[width=\textwidth]{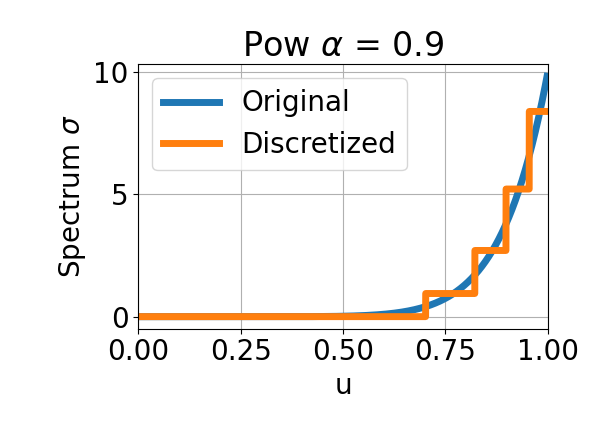}
    \end{subfigure}
    \hfill
    \begin{subfigure}[b]{0.301\textwidth}
        \centering
        \includegraphics[width=\textwidth]{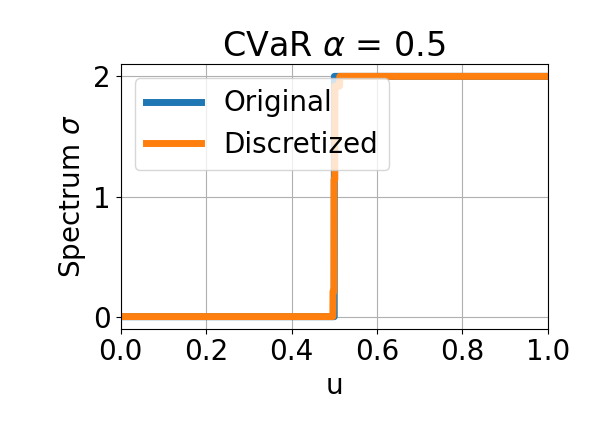}
    \end{subfigure}
    \hfill
    \begin{subfigure}[b]{0.301\textwidth}
        \centering
        \includegraphics[width=\textwidth]{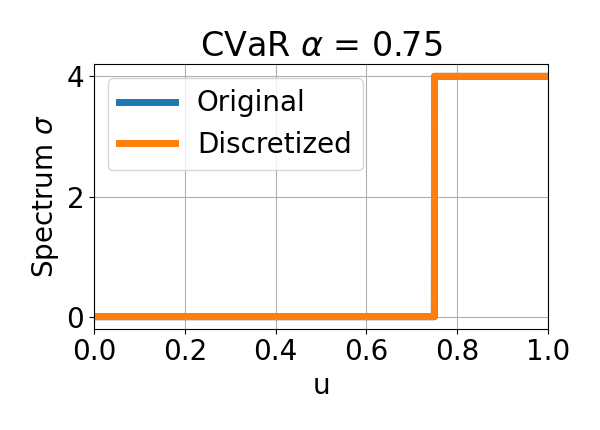}
    \end{subfigure}
    \hfill
    \begin{subfigure}[b]{0.301\textwidth}
        \centering
        \includegraphics[width=\textwidth]{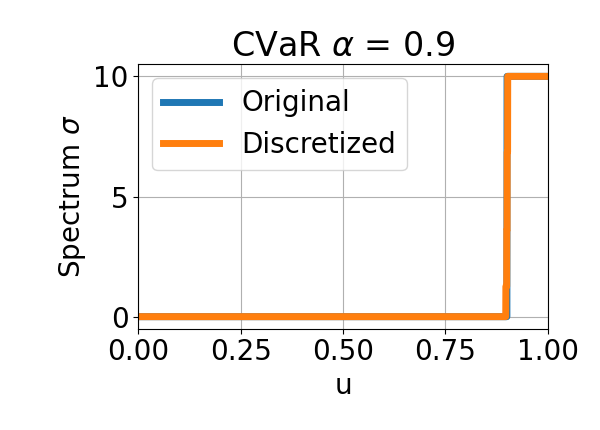}
    \end{subfigure}
    \caption{
    Visualization of the spectrum function of the Pow and CVaR risk measures and the discretized results.
    }
    \label{fig: other risk visualization}
\end{figure}

The discretization process can be implemented by minimizing (\ref{eq: approx problem}), and the results are shown in Figure \ref{fig: wang visualization}.
In this process, the number of discretizations $M$ is set to five.
Additionally, visualizations of other risk measures are provided in Figure \ref{fig: other risk visualization}, and the optimized values of the parameters for the discretized spectrum, which are introduced in (\ref{eq: discretized spectrum}), are reported in the Table \ref{tab: discretized results}.
Note that CVaR is already an element of the discretized class of spectral risk measures, as observed in Figure \ref{fig: other risk visualization}.
Therefore, CVaR can be precisely expressed using a parameter with $M=1$.

\begin{figure}[!thb]
    \centering
    \begin{subfigure}[b]{0.32\textwidth}
        \centering
        \includegraphics[width=\textwidth]{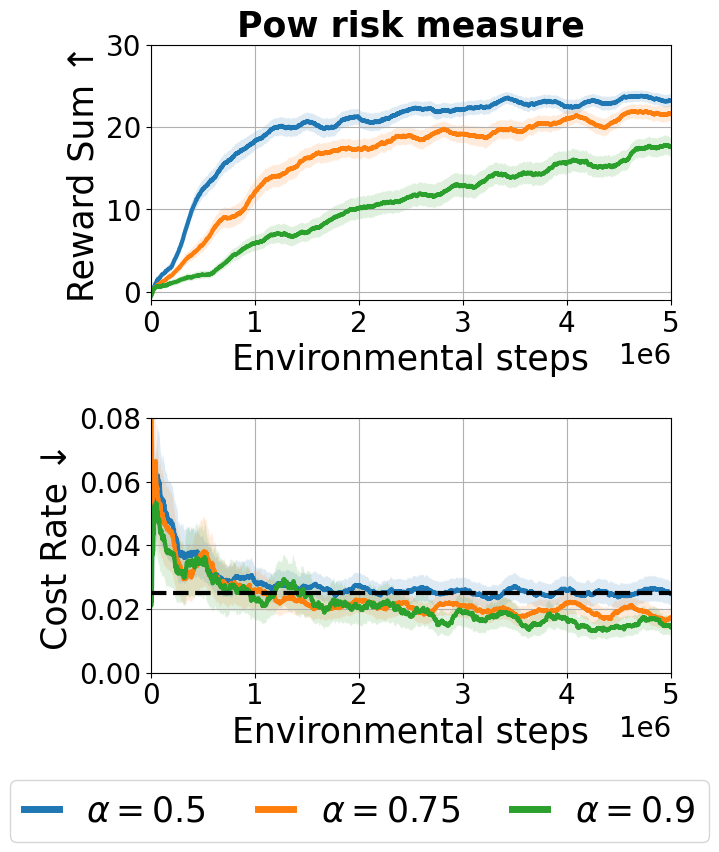}
    \end{subfigure}
    \hfill
    \begin{subfigure}[b]{0.32\textwidth}
        \centering
        \includegraphics[width=\textwidth]{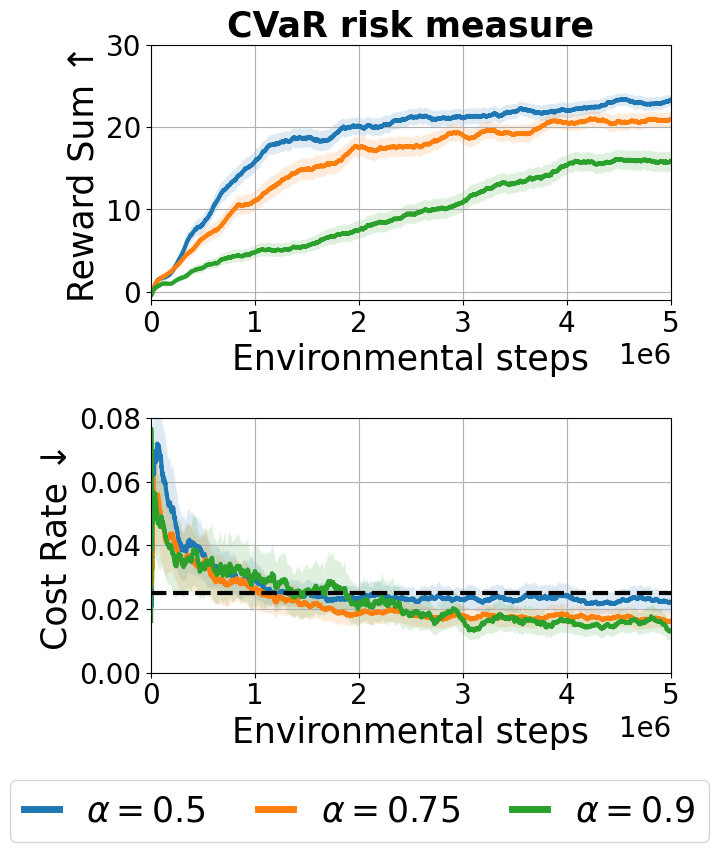}
    \end{subfigure}
    \hfill
    \begin{subfigure}[b]{0.32\textwidth}
        \centering
        \includegraphics[width=\textwidth]{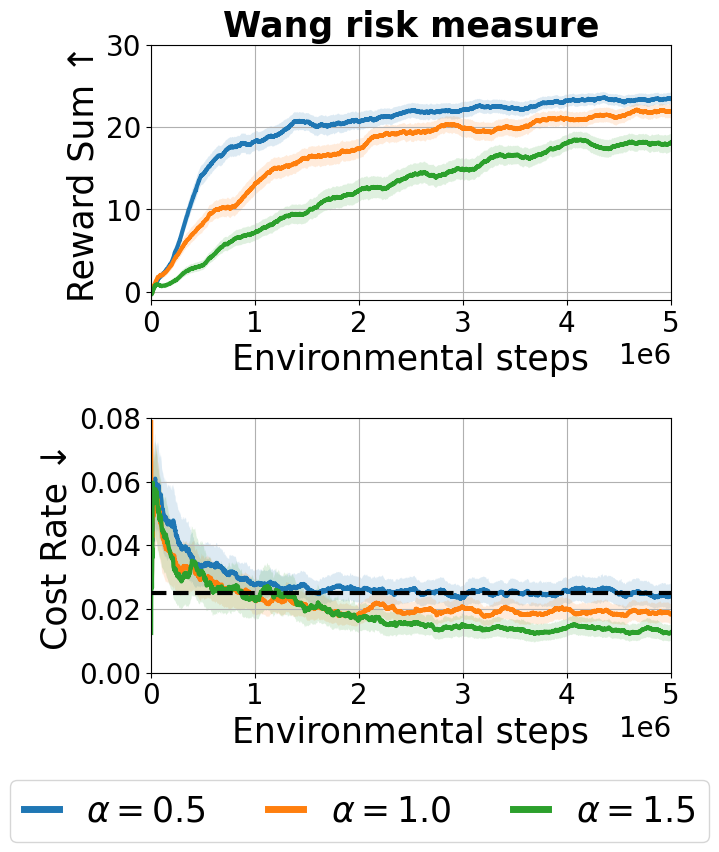}
    \end{subfigure}
    \caption{
    Training curves of the point goal task with different risk levels and risk measures.
    Each column shows the results on the risk measure whose name appears in the plot title.
    The solid line in each graph represents the average of each metric, and the shaded area indicates the standard deviation scaled by $0.2$. 
    The results are obtained by training algorithms with five random seeds.
    }
    \label{fig: training curves of other risk measures}
\end{figure}

Finally, we conducted experiments on the point goal task to check how the results are changed according to the risk level and the type of risk measures. 
The training curves for these experiments are shown in Figure \ref{fig: training curves of other risk measures}.

\begin{table}[h]
    \caption{Discretization results.}
    \label{tab: discretized results}
    \centering
    \resizebox{1.0\textwidth}{!}{%
    \begin{tabular}{c|c|c|c|c|c}
        \toprule
            &&  $\mathrm{Wang}_\alpha$&  && $\mathrm{Pow}_\alpha$\\
        \midrule
        $\alpha=0.5$ &$\{\eta_i\}_{i=1}^{M}$& $[0.515, 0.790, 1.091, 1.493, 2.191]$& $\alpha=0.5$ &$\{\eta_i\}_{i=1}^{M}$& $[0.200, 0.600, 1.000, 1.400, 1.800]$\\
          &$\{\alpha_i\}_{i=1}^{M-1}$& $[0.263, 0.541, 0.770, 0.926]$&  &$\{\alpha_i\}_{i=1}^{M-1}$&$[0.200, 0.400, 0.600, 0.800]$\\
        \midrule
        $\alpha=1.0$&$\{\eta_i\}_{i=1}^{M}$& $[0.294, 0.734, 1.417, 2.640, 5.517]$& $\alpha=0.75$&$\{\eta_i\}_{i=1}^{M}$& $[0.046, 0.574, 1.347, 2.308, 3.424]$\\
          &$\{\alpha_i\}_{i=1}^{M-1}$& $[0.409, 0.701, 0.878, 0.968]$&  &$\{\alpha_i\}_{i=1}^{M-1}$&$[0.417, 0.615, 0.765, 0.890]$\\
        \midrule
        $\alpha=0.5$ &$\{\eta_i\}_{i=1}^{M}$& $[0.180, 0.834, 2.253, 5.678, 16.419]$& $\alpha=0.9$&$\{\eta_i\}_{i=1}^{M}$& $[0.003, 0.947, 2.705, 5.216, 8.383]$\\
          &$\{\alpha_i\}_{i=1}^{M-1}$& $[0.579, 0.834, 0.945, 0.989]$&  &$\{\alpha_i\}_{i=1}^{M-1}$&$[0.701, 0.821, 0.898, 0.955]$\\
        \bottomrule
    \end{tabular}%
    }
\end{table}

\newpage

\end{document}